\documentclass[a4paper,12pt]{article}

\usepackage{amsmath}
\usepackage{amsthm}
\usepackage{amssymb}
\usepackage[authoryear,round]{natbib}
\usepackage{graphicx}
\usepackage[OT4]{fontenc}
\usepackage[utf8]{inputenc}
\usepackage{url}
\usepackage[unicode,debug,colorlinks=true,linkcolor=black,citecolor=black,urlcolor=black,pdfusetitle=true,breaklinks=true]{hyperref}
\newcommand{\toprule}{\hline}
\newcommand{\midrule}{\hline}
\newcommand{\botrule}{\hline}
\newcommand{\abs}[1]{\left| #1 \right|}

\newcommand{\okra}[1]{\left( #1 \right)}

\newcommand{\ceil}[1]{\left\lceil #1 \right\rceil}
\newcommand{\floor}[1]{\left\lfloor #1 \right\rfloor}
\newcommand{\kwad}[1]{\left[ #1 \right]}
\newcommand{\klam}[1]{\left\{ #1 \right\}}

\newcommand{\boole}[1]{{\bf 1}{\klam{#1}}}
\DeclareMathOperator{\card}{\#}
\DeclareMathOperator{\mean}{\mathbf{E}}
\DeclareMathOperator{\var}{Var}

\title{Corrections of Zipf's and Heaps' Laws \\ Derived from Hapax
  Rate Models}

\author{{\L}ukasz D\k{e}bowski%
  \\[1ex]
  \small{ORCID: 0000-0001-7136-5283}
  \\
  \small{Institute of Computer
    Science, Polish Academy of Sciences}
  \\
  \small{ul.\ Jana Kazimierza 5,
    01-248 Warszawa, Poland}
  \\
  \small{e-mail: ldebowsk@ipipan.waw.pl}}
\date{}

\begin{document}

\begin{titlepage}
\thispagestyle{empty}
  
% \layout

\begingroup
\let\newpage\relax

\null

\vspace{-6em}

\begin{center}
  \small This is the Author's Original Manuscript of an article
  published by Taylor \& Francis in Journal of Quantitative
  Linguistics on February 4, 2025, available at:
  \url{https://doi.org/10.1080/09296174.2025.2455764}
\end{center}

\vspace{-2em}

\maketitle
\endgroup

  \small\noindent \textbf{Abstract:} The article introduces
  corrections to Zipf's and Heaps' laws based on systematic models of
  the proportion of hapaxes, i.e., words that occur once.  The
  derivation rests on two assumptions. The first one is the standard
  urn model which predicts that marginal frequency distributions for
  shorter texts look as if word tokens were sampled blindly from a
  given longer text.  The second assumption posits that the hapax rate
  is a simple function of the text length.  Four such functions are
  discussed: the constant model, the cancelation model, the linear
  model, and the logistic model. As a simple illustration, it is shown
  that the logistic model yields the best fit for a sample of 14 texts
  in English. The need and the availability of more complex mixture
  models that reflect two-regime vocabularies for larger corpora is
  also discussed.
  \\[1ex]
  \\[1ex]
  \textbf{Prior versions:} This article is based on a presentation
  (abstract and slides) \emph{Principled Analytic Corrections of
    Zipf's Law} delivered at the Quantitative Linguistics Conference
  QUALICO 2023, Lausanne, Switzerland, June 28-30, 2023. That
  presentation was co-authored by Iv\'an Gonz\'alez Torre, who was
  about to contribute experimental data for the Catalan language but
  in the end he did not and he opted out from the full paper. The
  unrevised preprint of the submitted article is available at
  \url{https://arxiv.org/abs/2307.12896}.
  \\[1ex]
  \textbf{Funding:} This work received no funding from funding
  agencies.
  \\[1ex]
  \textbf{Disclosure statement:} There are no competing interests to
  declare.
  \\[1ex]
  \textbf{Acknowledgments:} I thank Iv\'an Gonz\'alez Torre, Antoni
  Hern\'andez-Fern\'andez, and Ji\v{r}\'\i{} Mili\v{c}ka for a
  fruitful discussion of the ideas contained in this paper. I am also
  grateful to the four anonymous reviewers for highly detailed and
  motivating reviews that helped to improve this paper considerably.

\end{titlepage}

\section{Introduction}
\label{secIntroduction}

%\citep{TuzziPopescuAltmann2010}

Zipf's law is the most famous and the oldest known statistical law
observed for texts in natural language.  It states that the $n$-th
most frequent word in a text appears approximately $n$ times less
often than the most frequent one.  This regularity was noticed by
\citet{Estoup16} for French and later recognized by \citet{Condon28}
and \citet{Zipf35} for English. Similar power-law distributions were
later observed in other domains of empirical research such as ecology,
sociology, economics, and physics \citep{Zipf49}. Hence Zipf's law is
often considered a hallmark of complex systems. The literature on
Zipf's law is vast but scattered over venues devoted to diverse
branches of science. Many older historical references can be found in
the online bibliography by \citet{LiWWW}. The departure point for a
modern statistically informed theory is provided by the monograph of
\citet{Baayen01}, which was influenced by a technical report by
\citet{Khmaladze88}. A study of Zipf's law across one hundred live
languages was reported by \citet{MehriJamaati17}.

In spite of the sheer size of existing literature, the question why
Zipf's law is so ubiquitous has not been fully answered.  Zipf's law
can be explained by qualitatively diverse mechanisms ranging from
disappointing monkey-typing models by \citet{Mandelbrot54} and
\citet{Miller57}, through preferential attachment by \citet{Simon55}
and game-theoretic considerations by \citet{HarremoesTopsoe05}, to
attempts of linking Zipf's law with semantics and information theory
\citep{Debowski21} as well as with deep learning and computable
deterministic sequences \citep{Debowski23}. There is a plethora of
explanations which look plausible but, upon closer scrutiny, ask for
deeper research.

This paper sets out a more modest goal. The aim is to derive a general
method for more systematic corrections to Zipf's and Heaps' laws for
texts of an arbitrary size. Zipf's law having been already introduced,
Heaps' law is a quantitative linguistic law that predicts the total
number of distinct words as a function of the text length
\citep{Herdan64,Heaps78}.  Our derivation of corrections stems from
two modeling assumptions:
\begin{itemize}
\item The first assumption is the standard urn model
  \citep{Khmaladze88,Baayen01} which states that the marginal word
  frequency distributions for shorter texts look as if word tokens
  were sampled blindly from a longer text.  In the limit of an
  unbounded source, the word distributions look as if texts were
  generated by a memoryless stochastic process. The urn model yields a
  surprisingly good prediction of the marginal distributions although
  it obviously fails at predicting statistical dependencies among
  consecutive words.
\item Our second assumption is that the hapax rate, i.e., the
  proportion of words appearing once, can be reasonably approximated
  by a simple analytic function of the text size. We will show that a
  simple parametric model for the hapax rate is enough to derive the
  type-token and rank-frequency plots that look more plausible than
  more popular but theoretically less informed models.  In particular,
  a one-parameter model is sufficient for texts of a size of a novel,
  where the hapax rate has a decaying shape. By contrast, adding one
  more parameter can account for larger corpora of texts, where the
  hapax rate is $U$-shaped \citep{Fan10} and a second regime of Zipf's
  law emerges \citep{FerrerSole01b, MontemurroZanette02,
    PetersenOthers12, GerlachAltmann13, WilliamsOthers15}.
\end{itemize}

The statistical theory developed in this article draws from works by
\citet{Khmaladze88} and \citet{Baayen01} with some later add-ons by
\citet{Milicka09,Milicka13}, \citet{GerlachAltmann13},
\citet{FontClosCorral15}, and \citet{Davis18}, which were likely
developed independently. In our contribution, we systematically review
and consolidate a collection of various mathematical formulae and,
most importantly, we identify the fundamental role of the hapax rate
function in deriving the type-token and rank-frequency relationships
that correct Heaps' and Zipf's laws. Consequently, we offer a
preliminary test on a sample of 14 texts in English that informs how
all these theoretically derived formulae fit linguistic reality. It
turns out that if we manage to model the hapax rate well enough then
the standard urn model predicts word frequency distributions
surprisingly precisely. This improved fitting should not be taken as a
goal by itself but rather as a practical means to remove systematic
errors so as to pave a road for improved testing of theories of a
linguistic importance in the future.

The organization of the paper is as follows. 
%\ref{secIntroduction}
In \S \ref{secPreliminaries} we present preliminaries: \S
\ref{secBasic} introduces the basic concepts in the study of word
frequency distributions, whereas \S \ref{secPerspective} describes
fundamental challenges in modeling Zipf's law. In \S \ref{secGeneral},
we expose the standard theory of the urn model and a relatively less
developed theory of the analytic vocabulary size function: \S
\ref{secUrn} deals with the finite urn model, \S \ref{secMemoryless}
concerns the memoryless source, \S \ref{secConvergence} treats the
stochastic convergence, whereas \S \ref{secAnalytic} develops the
theory of analytic vocabulary growth and hapax rate. In \S
\ref{secModels}, we introduce four parametric models for the hapax
rate: the constant model in \S \ref{secConstant}, the cancelation
model in \S \ref{secDavis}, the linear model in \S \ref{secLinear},
and the logistic model in \S \ref{secLogistic}. In \S
\ref{secExperiments}, we test these models experimentally: \S
\ref{secData} introduces the data, \S \ref{secMethods} describes our
methods, \S \ref{secResults} summarizes the results, whereas \S
\ref{secDiscussion} offers the discussion. In \S \ref{secMixture}, we
discuss possible extensions of hapax rate models for future research.
The article is concluded in \S \ref{secConclusion}.

\section{Preliminaries}
\label{secPreliminaries}

In this section, we introduce the basic concepts in the study of word
frequency distributions and we develop a preliminary discussion of
Zipf's and Heaps' laws. We begin with some necessary definitions and
then we explain why the literal Zipf law can be only a theoretical
idealization.

\subsection{Basic entities}
\label{secBasic}

Suppose that we count certain objects in empirical data.  We assume
that these objects come in many shapes but we are able to tell that
some shapes are identical or similar enough to be treated as instances
of the same type.  To fix our attention, let us call these objects
words and let the empirical data be a text, i.e., a fixed sequence of
words. For each word $w$ and a text being a fixed sequence of words
$\mathbf{t}=(t_1,t_2,...,t_n)$, we define the frequency of word $w$ as
\begin{align}
  f(w):=\sum_{i=1}^n \boole{t_i=w},
\end{align}
where $\boole{\text{true}}:=1$ and $\boole{\text{false}}:=0$.  The
vocabulary of text $\mathbf{t}$ is the set of words that appear in
this text, namely,
\begin{align}
  \mathbf{w}:=\klam{w: f(w)>0}.
\end{align}
  
To distinguish two meanings of the word ``word'', the elements of text
$\mathbf{t}$ will be called \emph{tokens}, whereas the elements of
vocabulary $\mathbf{w}$ will be called \emph{types}. The number of
tokens is denoted $n$, whereas the number of types is denoted $v$. We
may write
\begin{align}
  n&=\sum_{w\in\mathbf{w}} f(w),
  &
  v&=\sum_{w\in\mathbf{w}} 1.                                 
\end{align}
We also observe that the number of tokens is the length of the text
$n=\abs{t}$ and the number of types is the cardinality of the
vocabulary $v=\card \mathbf{w}$.  We consistently use lower-case
symbols to denote fixed objects, reserving upper-case ones for random
variables.

Let us proceed to the next important idea, namely, the notion of
ranks.  Let us introduce a total order on types
$w_1,w_2,...,w_v\in\mathbf{w}$, where $w_i\neq w_j$ for $i\neq j$, by
sorting them according to frequencies:
\begin{align}
f(w_1)\ge f(w_2)\ge ... \ge f(w_v). 
\end{align}
We define the \emph{rank} of a word as
\begin{align}
r(w_k):=k.
\end{align}
That is, the most frequent word $w_1$ has rank $1$, the second most
frequent word $w_2$ has rank $2$, and so on.  We can also define the
\emph{frequency function} as
\begin{align}
  f_r:=f(w_r).
\end{align}
We may express $n=\sum_{r=1}^v f_r$.

The frequency function is somewhat inconvenient to evaluate. It is
simpler to evaluate a related entity, namely, the maximal rank with a
given frequency,
\begin{align}
  \label{RankFunction}
 r_k:=\max\klam{r: f_r\ge k}=\sum_{w\in\mathbf{w}} \boole{f(w)\ge k}. 
\end{align}
Function $k\mapsto r_k$ will be called the \emph{rank function}.

Subsequently, we will define the frequency spectrum, starting with the
sets of types that appear exactly $k$ times,
\begin{align}
  \mathbf{w}_k:=\klam{w: f(w)=k}.
\end{align}
Elements of $\mathbf{w}_1$, i.e., types that appear exactly once, are
called \emph{hapax legomena}, or succinctly \emph{hapaxes}. The number
of elements of $\mathbf{w}_k$ is denoted $v_k:=\card \mathbf{w}_k$. In
particular, $v_1$ is the number of hapaxes. We may express
\begin{align}
  v&=\sum_{k=1}^\infty v_k,  & n&=\sum_{k=1}^\infty k v_k.
\end{align}
Sequence $(v_1,v_2,...)$ is called the \emph{frequency spectrum}. The
spectrum elements $v_k$ are sometimes called the frequencies of
frequencies.

Can we derive the rank function $k\mapsto r_k$ from the frequency
spectrum?  Indeed, we have $v_k=r_k-r_{k+1}$ and $r_1=v$, so
\begin{align}
  \label{RankFunctionV}
  r_f=v-\sum_{k=1}^{f-1} v_k.
\end{align}
This linear formula conveniently allows to compute the expectation
of the rank function when we consider random texts. Hence, we would
like to advocate that the rank function $f\mapsto r_f$ is a simpler
behaved object than the frequency function $r\mapsto f_r$.

\subsection{Zipf's law in perspective}
\label{secPerspective}

By the definition of ranks, the frequency function $r\mapsto f_r$ is
decreasing.  Shall we expect a particular form of this function?
Observe that
\begin{align}
  kr_k=k\sum_{w\in\mathbf{w}} \boole{f(w)\ge k}
  \le
  \sum_{w\in\mathbf{w}} f(w)=n, 
\end{align}
by the Markov inequality $k\boole{y\ge k}\le y$ for $y\ge 0$, which is
frequently used in probability calculus. Hence we obtain a
\emph{harmonic bound} for the frequency and rank functions
\begin{align}
  f_r&\le\frac{n}{r}, & r_f&\le\frac{n}{f}.
\end{align}

The typical shape of the frequency function for orthographic words in
an average text written in natural language is relatively close to this
upper bound. Namely, it turns out that function $r\mapsto f_r$ is not
only decreasing but equals approximately
\begin{align}
f_r\approx C/r,
\end{align}
where $C$ is a constant.  This empirical relationship is called
\emph{Zipf's law}. It was discovered by \citet{Estoup16} and
\citet{Condon28} and popularized by \citet{Zipf35,Zipf49}.

In the following reasoning, we will interpret Zipf's law quite
literally. We will assume that
\begin{align}
  \label{Zipf}
  f_r=\floor{\frac{v}{r}}
\end{align}
and we will inspect mathematical consequences of formula
(\ref{Zipf}). We use the floor function
$x\mapsto\floor{x}:=\max\klam{k\in\mathbb{Z}:k\le x}$ since
frequencies are non-negative integers.

Let $k\mapsto r_k$ be the rank function given by
(\ref{RankFunction}). By (\ref{Zipf}), we have $r_k\approx
v/k$. Hence,
\begin{align}
  \label{Lotka}
  v_k=r_k-r_{k+1}\approx
  \frac{v}{k}-\frac{v}{k+1}=\frac{v}{k(k+1)}.
\end{align}
Formula (\ref{Lotka}) is called \emph{Lotka's law} \citep{Lotka26}. It
is approximately correct for $v_k>1$ assuming the exact Zipf law
(\ref{Zipf}). In particular, by (\ref{Lotka}), the number of hapaxes
approximately equals half the total number of types,
\begin{align}
 v_1\approx \frac{v}{2},
\end{align}
which is pretty large. In reality, the \emph{hapax rate} $v_1/v$
differs significantly from $1/2$ \citep{Baayen01,Fan10}, which will be
the base for our amendments in \S \ref{secModels}.

Let frequency $f_{\bullet}$ be the minimal frequency $f$ such that
$v_f=1$ and let $r_{\bullet}=r_{f_{\bullet}}$ be the corresponding
rank.  By Lotka's law (\ref{Lotka}), we may approximate
$f_{\bullet}\approx \sqrt{v}$ and by Zipf's law (\ref{Zipf}), we may
put $r_{\bullet}\approx \sqrt{v}$.  Ignoring the subtlety that
$\sqrt{v}$ is not an integer, we may compute
\begin{align}
  n
  &\approx
  \sum_{r=1}^{r_{\bullet}} f_r + \sum_{f=1}^{f_{\bullet}} f v_f
  \approx
  \sum_{r=1}^{\sqrt{v}} \frac{v}{r}+\sum_{f=1}^{\sqrt{v}}
    \frac{v}{f+1}
    \nonumber\\
  &\approx
    2\sum_{r=1}^{\sqrt{v}}\frac{v}{r}-\frac{v}{1}
  \approx
    2v(\log\sqrt{v}+\gamma)-v
  =
    v(\log v+2\gamma-1),
  \label{TokensZipf}
\end{align}
where $\log x$ is the natural logarithm of $x$ and
\begin{align}
  \lim_{k\to\infty}\kwad{\sum_{r=1}^k\frac{1}{r}-\log k}=\gamma,
\end{align}
where Euler's gamma is $\gamma\approx 0.577$.  In particular, assuming
the literal Zipf law (\ref{Zipf}), the relative frequency of the most
frequent type is
\begin{align}
  \label{RelativeFrequency}
  \frac{f_1}{n}\approx \frac{1}{\log v+2\gamma-1},
\end{align}
which tends to zero for the number of types $v$ tending to infinity.

Since, for a given language, the most frequent word is usually a fixed
functional word, the relative frequency of the most frequent word
$f_1/n$ is quite stable and it does not depend strongly on a
particular text. Hence relationship (\ref{RelativeFrequency}) suggests
some corrections to Zipf's law depending on the text size.
\citet{Orlov82}, followed by \citet{Baayen01} and \citet{Davis18},
supposed that Zipf's law (\ref{Zipf}) holds exactly only for a text of
a certain length.

In the following example, let $n$ be this specific length. Let us
assume that $f_1/n\approx 0.1$, which holds approximately for English
\citep{Shannon51}. From (\ref{RelativeFrequency}), we may estimate the
corresponding number of types
\begin{align}
  \label{TypesZipf}
  v\approx \exp\okra{\frac{n}{f_1}-2\gamma+1}\approx 18\,883.
\end{align}
Consequently, by estimate (\ref{TokensZipf}), we obtain
$n\approx 83\,653$ tokens, which is somewhat less than the average
length of a novel.  For both shorter and longer texts we should expect
deviations.

To account for these deviations, various heuristic corrections driven
by empirical observations were proposed. For example, a correction
proposed by \citet{Mandelbrot54} reads
\begin{align}
  \label{Mandelbrot}
  f_r\approx \floor{\frac{B+v}{B+r}}^\alpha, \quad \alpha>1.
\end{align}
In particular, \citet{Debowski02c} combined Mandelbrot's correction
with Orlov's idea by adjusting parameters $\alpha$ and $B$ so that
$f_1/n$ be constant. The linguistic reality is somewhat different,
however. \citet{FerrerSole01b} discovered that the empirical data for
large collections of texts are closer to two regimes: Zipf's law
(\ref{Zipf}) for small ranks and Mandelbrot's correction
(\ref{Mandelbrot}) with $\alpha\gg 1$ for large
ranks. \citet{MontemurroZanette02} observed that the second regime can
be sometimes closer to an exponential decay. The shape of the
frequency function tail depends on the composition of the collection
of texts, being steeper for a single-author collection and decaying
slower for a mixture of texts by many authors.

The heuristic Mandelbrot correction predicts a power-law growth of the
number of types. In fact, if Mandelbrot's correction
(\ref{Mandelbrot}) holds exactly and the proportion of the most
frequent type is constant, $f_1/n = p_1$, then we have
\begin{align}
 np_1=f_1= \okra{\frac{B+v}{B+1}}^\alpha.
\end{align}
Hence we obtain \emph{Herdan-Heaps' law}
\begin{align}
  \label{HerdanHeaps}
  v= (B+1)n^{1/\alpha}p_1^{1/\alpha}-B\approx Cn^{1/\alpha},
\end{align}
proposed likely independently by \citet{Herdan64} and \citet{Heaps78}.
This derivation is approximate and prone to gross errors since the
best fit of the power-law type-token relationship $v\approx C n^\beta$
yields usually $\beta\approx 0.8$ \citep{Kornai02} or even
$\beta\approx 0.6$ for spoken texts
\citep{HernandezFenrandezOthers19,TorreOthers19}, whereas $1/\alpha$
estimated by fitting Mandelbrot's correction (\ref{Mandelbrot}) is
often closer to $1$.  Moreover, the number of hapaxes predicted by
Mandelbrot's correction is much larger than it follows from
Herdan-Heaps' law (\ref{HerdanHeaps}) via the urn model, as we will
discus in \S \ref{secConstant}.

\section{General theory}
\label{secGeneral}

In the following, we will present a more precise account of word
frequency distributions. We will expose the standard theory of the urn
model and a less developed theory of the analytic vocabulary size
function. These two theories describe the expected frequency spectrum
in function of the text length.  The content of this section is formed
as a kind of a review of works \citep{Khmaladze88, Baayen01,
  Milicka09, Milicka13, GerlachAltmann13, FontClosCorral15, Davis18}
with some extensions of ours. We try to present a concise but
systematic exposition.

\subsection{Urn model}
\label{secUrn}

In the context of quantitative linguistic research, the standard urn
model is due to \citet{Khmaladze88} and \citet{Baayen01}.  Up to
varying degrees, it was probably independently rediscovered by
\citet{Milicka09,Milicka13}, \citet{GerlachAltmann13},
\citet{FontClosCorral15}, and \citet{Davis18}.  The idea of this
framework is that word frequency distributions for a given part of a
finite text look as if tokens were selected blindly without
replacement from an urn that contains all tokens from the whole
text. That is, if we are interested in the marginal distribution of
types, we can ignore the sequential order of tokens.

Formally, suppose that we know the exact shape of the rank function
for a certain text $\mathbf{t}^*=(t^*_1,t^*_2,...,t^*_{n^*})$. We can
derive the expected values of the rank function for the random text of
length $n<n^*$ that is sampled without replacement from population
$\mathbf{t}^*$.  We introduce random variable being text
$\mathbf{T}=(T_1,T_2,...,T_{n})$ with random tokens $T_i$ that are
sampled from $\mathbf{t}^*$.  The probability distribution of
$\mathbf{T}$ is
\begin{align}
  P(\mathbf{T}=\mathbf{t})
  &=
    \frac{\prod_{w} f^*(w)!/[f^*(w)-f(w)]!}{n^*!/(n^*-n)!},
\end{align}
where $f^*(w)$ and $f(w)$ are the frequencies of word $w$ in texts
$\mathbf{t}^*$ and $\mathbf{t}$ respectively.

Now let us consider the sequence of binary indicators
$\mathbf{B}=(B_1,B_2,...,B_{n})$ where $B_i:=\boole{T_i=w}$ for a
fixed word $w$.  For brevity, let $f^*:=f^*(w)$ and $f:=f(w)$. We
derive
\begin{align}
  P(\mathbf{B}=\mathbf{b})
  &=
    \frac{[n^*-f^*]!/[n^*-n-f^*+f]!\cdot f^*!/[f^*-f]!}{n^*!/(n^*-n)!}
   % \nonumber\\
  % &
    =\frac{\binom{n^*-n}{f^*-f}}{\binom{n^*}{f^*}}
    ,
\end{align}
where the binomial coefficient is denoted
\begin{align}
  \binom{n}{k}:=\frac{n!}{k!(n-k)!}.
\end{align}
Let us denote the frequency of word $w$ in text $\mathbf{T}$ as
$F:=F(w)$.  The number of distinct sequences $\mathbf{b}$ that induce
a given frequency $f$ is $\binom{n}{f}$.  Hence we evaluate that the
frequency of the word is dictated by the hypergeometric distribution
\begin{align}
  P(F=f)
  &=
    \frac{\binom{n}{f}\binom{n^*-n}{f^*-f}}{\binom{n^*}{f^*}}
    .
\end{align}

Let $V$ be the number of types and $V_k$ be the spectrum elements for
text $\mathbf{T}$. We have
\begin{align}
  \label{V}
  V&=\sum_{w} \boole{F(w)>0},
  \\
  \label{Vk}
    V_k&=\sum_{w} \boole{F(w)=k}.                                 
\end{align}
Hence we derive the expectations
\begin{align}
  \label{ExpV}
  \mean V&=\sum_{w} P(F(w)>0)=
           v^* - \sum_{k^*=1}^\infty v^*_{k^*}
           \frac{\binom{n^*-n}{k^*}}{\binom{n^*}{k^*}},
  \\
  \label{ExpVk}
  \mean V_k&=\sum_{w} P(F(w)=k)=
             \sum_{k^*=k}^\infty v^*_{k^*} 
             \frac{\binom{n}{k}\binom{n^*-n}{k^*-k}}{\binom{n^*}{k^*}},
\end{align}
where $v^*_k$ are the spectrum elements for text $\mathbf{t}^*$ and we
use the facts that $P(F(w)>0)=1-P(F(w)=0)$ and
$\binom{n}{0}=1$. Expectations $\mean V$ and $\mean V_k$ are close to
the empirical values for natural language if we use observed values
$v^*_k$ \citep{Milicka09,Milicka13}.

Similarly, we derive the expected rank function. Let $R_f$ be the rank
function for text $\mathbf{T}$, respectively. Using formula
(\ref{RankFunctionV}), the Chu-Vandermonde identity that expresses the
normalization of the hypergeometric distribution
\begin{align}
  \sum_{k=0}^{k^*} \frac{\binom{n}{k}\binom{n^*-n}{k^*-k}}{\binom{n^*}{k^*}}=1,
\end{align}
and putting $v_0:=0$, we obtain the expected rank function
\begin{align}
  \mean R_f
  =\mean V-\sum_{k=1}^{f-1} \mean V_{k}
  &=v^*-\sum_{k=0}^{f-1} \sum_{k^*=k}^\infty v^*_{k^*}
    \frac{\binom{n}{k}\binom{n^*-n}{k^*-k}}{\binom{n^*}{k^*}}
    \nonumber\\
  &=r^*_f -\sum_{k^*=f}^\infty v_{k^*} \sum_{k=0}^{f-1}
    \frac{\binom{n}{k}\binom{n^*-n}{k^*-k}}{\binom{n^*}{k^*}}
    ,
    \label{ExpectedRanks}
\end{align}
% $a\wedge b:= \min(a,b)$.
Hence, to obtain the expectation of $R_k$, we have to subtract from
$r^*_k$ the number of types appearing $\ge k$ times multiplied by the
probability of observing a given type $<k$ times.

\subsection{Memoryless source}
\label{secMemoryless}

This setting was also studied by \citet{Khmaladze88} and
\citet{Baayen01}. If we let the size of the urn tend to infinity,
preserving the proportion of tokens of each type then we obtain a
memoryless source. The relative frequencies tend to probabilities and
sampling without replacement becomes indistinguishable from sampling
with replacement. Each draw from the urn becomes a probabilistically
independent copy of previous draws.

Formally, we are given a certain probability distribution of word
types $w\mapsto p(w)$, where $p(w)\in[0,1]$ and $\sum_w p(w)=1$. The
theoretical vocabulary, i.e., the domain of function $w\mapsto p(w)$
may be countably infinite.  We consider a memoryless source which
emits sequences of words according to the probability distribution
$w\mapsto p(w)$.  We introduce random text
$\mathbf{T}=(T_1,T_2,...,T_{n})$ with random tokens $T_i$.  The
probability distribution of text $\mathbf{T}$, meant as a fixed
sequence of tokens, is simply
\begin{align}
  P(\mathbf{T}=\mathbf{t})
  =
    \prod_{w}
    p(w)^{f(w)},
\end{align}
where $f(w)$ is the frequency of word $w$ in text
$\mathbf{t}$. Defining the sequence of binary indicators
$\mathbf{B}=(B_1,B_2,...,B_{n})$ where $B_i:=\boole{T_i=w}$ for a
fixed word $w$ and putting $f:=f(w)$ and $p:=p(w)$, we derive
\begin{align}
  P(\mathbf{B}=\mathbf{b})
  =
    p^{f}(1-p)^{n-f}.
\end{align}
Let us denote the frequency of word $w$ in text $\mathbf{T}$ as
$F:=F(w)$. Since the number of distinct sequences $\mathbf{b}$ that
induce a given frequency $f$ is $\binom{n}{f}$, the frequency of the
word is dictated by the binomial distribution
\begin{align}
  P(F=f)
  &=
    \binom{n}{f}\, p^{f}(1-p)^{n-f}.
\end{align}

Let $V$ be the number of types in text $\mathbf{T}$ and let $V_l$ be
the number of types with frequency $l$ in text $\mathbf{T}$.  By
(\ref{V})--(\ref{Vk}), we derive the expectations
\begin{align}
  \label{ExpVS}
  \mean V&=\sum_w [1-(1-p(w))^n]
           \approx \sum_w [1-e^{-np(w)}],
  \\
  \label{ExpVkS}
  \mean V_k&=\sum_w \binom{n}{k} p(w)^{k}(1-p(w))^{n-k}
             \approx \sum_w \frac{[np(w)]^k}{k!} e^{-np(w)},
\end{align}
where the approximations are valid for $n\gg 1$ and a finite potential
vocabulary by the asymptotic convergence of the binomial distribution
to the Poisson distribution.  Let $R_f$ be the rank function for text
$\mathbf{T}$, respectively.  Using formula (\ref{RankFunctionV}), we
obtain
\begin{align}
  \mean R_f
  =\mean V-\sum_{k=1}^{f-1} \mean V_{k}
    &=\sum_{w} \kwad{1-\sum_{k=0}^{f-1}
    \binom{n}{k} p(w)^{k}(1-p(w))^{n-k}}
    \nonumber\\
    &\approx \sum_w \kwad{1-\sum_{k=0}^{f-1}
    \frac{[np(w)]^k}{k!}e^{-np(w)}}.
\end{align}

\subsection{Probabilistic convergence}
\label{secConvergence}

Let us consider the memoryless source as in the previous section. We
will briefly comment on the probabilistic convergence for this
model. For this aim, we introduce the absolute total order on types
$w_1,w_2,w_3,...$, where $w_i\neq w_j$ for $i\neq j$, by sorting them
according to probabilities:
\begin{align}
 p(w_1)\ge p(w_2)\ge p(w_3)\ge ...\, .
\end{align}
Consequently, we define the \emph{theoretical rank} of a word defined as
\begin{align}
  r(w_k):=k.
\end{align}
We also define the \emph{theoretical probability function} and the
\emph{theoretical rank function},
\begin{align}
  p_r&:=p(w_r),
  &
    s_t&:=\max\klam{r: p_r\ge t}. 
\end{align}
Observe that $ts_t, rp_r\le\sum_{i=1}^\infty p(w_i)=1$. Hence, we
obtain a \emph{harmonic bound} for the theoretical probability and rank
functions, namely,
\begin{align}
  p_r&\le \frac{1}{r},
  &
    s_t&\le \frac{1}{t}.
\end{align}

Analogously, for the random text $\mathbf{T}=(T_1,T_2,...,T_{n})$, we
introduce a random total order on types $W_1,W_2,W_3,...$, where
$W_i\neq W_j$ for $i\neq j$, by sorting them according to the
empirical frequencies:
\begin{align}
 F(W_1)\ge F(W_2)\ge F(W_3)\ge ...\, . 
\end{align}
Consequently, we define the \emph{empirical rank} of a word defined as
\begin{align}
  R(W_k):=k.
\end{align}
We also define the \emph{empirical frequency function} and the
\emph{empirical rank function},
\begin{align}
  F_r&:=F(W_r)
  \\
  R_t&:=\max\klam{r: F_r\ge t}=R_{\ceil{t}}. 
\end{align}

For the memoryless source, we have $\mean F(w)=np(w)$ and
$\var F(w)=np(w)(1-p(w))$. By the strong law of large numbers,
relative empirical frequencies converge to probabilities,
\begin{align}
  \lim_{n\to\infty} \frac{F(w)}{n}=p(w) \text{ almost surely}.
\end{align}
If $p(w_1)> p(w_2)> p(w_3)> ...$ then the empirical ranking converges
to the theoretical one, 
\begin{align}
  \lim_{n\to\infty} W_k&=w_k  \text{ almost surely},
  \\
  \lim_{n\to\infty} R(w)&=r(w) \text{ almost surely}.
\end{align}
Consequently, the empirical relative frequency function converges to
the theoretical probability function,
\begin{align}
  \label{SLLN}
  \lim_{n\to\infty} \frac{F_r}{n}&=p_r \text{ almost surely},
  \\
  \lim_{n\to\infty} \frac{\mean F_r}{n}&=p_r,
\end{align}
where the second equality follows by the Lebesgue dominated
convergence since $F_r/n\le 1$. We note that convergence (\ref{SLLN})
can be quite slow, though.

The pointwise convergence (\ref{SLLN}) implies the analogous
convergence for the rank function,
\begin{align}
\lim_{n\to\infty} R_{nt}=s_t \text{ almost surely},
\end{align}
since
\begin{align}
  R_{nt}=\max\klam{r: \frac{F_r}{n}\ge t}\xrightarrow[n\to\infty]{}
  \max\klam{r: p_r\ge t}=s_t.
\end{align}
Moreover, for an arbitrary size of the theoretical vocabulary, by the
Markov inequality, we notice an upper bound
\begin{align}
 R_k=\sum_w\boole{F(w)\ge k}\le \sum_w \frac{F(w)}{k}.
\end{align}
Since $\sum_w F(w)=n$, we obtain a uniform bound $R_{nt}\le 1/t$.
Thus, also the convergence in expectation,
\begin{align}
  \lim_{n\to\infty} \mean R_{nt}=s_t,
\end{align}
holds by the Lebesgue dominated convergence.

\subsection{Analytic functions}
\label{secAnalytic}

The idea of analytic functions in the study of word frequency
distributions is due to \citet{Baayen01}. It was later likely
independently discovered by \citet{Davis18}.  Suppose that the
potential vocabulary is finite and the texts are long, that is,
$n\gg 1$. Then for the memoryless source, we may approximate formulae
(\ref{ExpVS})--(\ref{ExpVkS}) as
\begin{align}
  \mean V&\approx g(n):= \sum_w [1-e^{-np(w)}],
  \\
  \mean V_k&\approx g(n|k):= \sum_w \frac{[np(w)]^k}{k!} e^{-np(w)},
\end{align}
where series $g(n)$ and $g(n|k)$ are analytic functions of a real
argument $n\ge 0$. In the following, we will consider functions $g(n)$
and $g(n|k)$ as convenient but not fully rigorously justified
approximations of expectations $\mean V$ and $\mean V_k$ for an
infinite potential vocabulary and any text length $n$, including any
real argument $n\ge 0$.

Function $g(n)$ that approximates the expected number of types for a
given text size $n$ will be called the \emph{smoothed vocabulary size}
or the \emph{smoothed number of types}.  Similarly, functions
$g(n|k)$, called the \emph{smoothed frequency spectrum elements},
approximate the expected frequency spectrum. The first smoothed
spectrum element $g(n|1)$ is also called the \emph{smoothed number of
  hapaxes}.  It was explicitly observed by \citet[Definition
2.11]{Baayen01} for $k=1$ and probably independently by
\citet{Davis18} for any $k\ge 1$ that the smoothed frequency spectrum
elements $g(n|k)$ can be expressed via the consecutive derivatives of
smoothed number of types $g(n)$,
\begin{align}
  g(n|k)=-\frac{(-n)^{k}}{k!} g^{(k)}(n).
  \label{Derivatives}
\end{align}
We may also approximate the expected rank function as
\begin{align}
  \mean R_f
  =\mean V-\sum_{k=1}^{f-1} \mean V_k
  \approx
  g(n||f)
  &:=g(n)-\sum_{k=1}^{f-1} g(n|k)
  \nonumber\\
  &=g(n)+\sum_{k=1}^{f-1} \frac{(-n)^{k}}{k!} g^{(k)}(n),
  \label{Taylor}
\end{align}
where function $g(n||f)$ is called the \emph{smoothed rank
  function}. Formula (\ref{Taylor}) can be easily remembered as the
truncated Taylor series for $g(0)$ expanded around point $n$. The
take-away is that if we can guess a certain analytic smoothed
vocabulary size function $g(n)$ then by taking the Taylor series
thereof, we may evaluate also the smoothed rank function $g(n||f)$ for
any text size $n$. The necessary conditions for the smoothed
vocabulary size are
\begin{align}
  g(1)&=1, &
  g(n|k)&\ge 0
  \text{ for }
  k\ge 1, &
  g(0)&=g(n)-\sum_{k=1}^\infty g(n|k)=0.
  \label{GConditions}
\end{align}

The smoothed vocabulary size and the smoothed frequency spectrum can
be expanded as the Taylor series also around an arbitrarily chosen
point $n^*$. We obtain
\begin{align}
  \label{GN}
  g(n)&=g(n^*)-\sum_{k^*=1}^\infty g(n^*|k^*) \okra{1-\frac{n}{n^*}}^{k^*},
  \\
  \label{GNk}
  g(n|k)&=\sum_{k^*=k}^\infty \binom{k^*}{k} g(n^*|k^*)
  \okra{\frac{n}{n^*}}^{k}\okra{1-\frac{n}{n^*}}^{k^*-k}.
\end{align}
These formulae were observed by \citet{Davis18}, who derived them as
approximations of equations (\ref{ExpV})--(\ref{ExpVk}) for the finite
urn model. In fact, to obtain formulae (\ref{GN})--(\ref{GNk}), it
suffices to approximate $v^*\approx g(n^*)$,
$v^*_{k^*}\approx g(n^*|k^*)$, and
\begin{align}
 \binom{r}{k}\approx \frac{r^k}{k!},
\end{align}
which is valid for $n^*,n\gg k^*,k$.

Let us make a series of useful observations, which seem new. These
arise from investigating the rate of hapaxes. We may define the
\emph{smoothed hapax rate} $h(u)$ as the proportion of the smoothed
number of hapaxes to the smoothed number of types,
\begin{align}
  h(\log n):=\frac{g(n|1)}{g(n)}
  \approx 
  \frac{\mean V_{1}}{\mean V}
  .
\end{align}
For $n=0$, the above formula should be read as condition
\begin{align}
  \lim_{u\to-\infty} h(u)=\frac{g(0|1)}{g(0)}.
\end{align}
Similar limit transitions for $n=0$ are tacitly adopted in formulae
that appear later in this article.

Variable $u=\log n$ is a natural choice of the argument for the
smoothed hapax rate for the following reason.  Namely, we observe that
the smoothed hapax rate $h(u)$ carries the same information as the
smoothed vocabulary size $g(n)$ since there is a one-to-one
correspondence between $h(u)$ and $g(n)$.  It is so since solving the
differential equation
\begin{align}
  n \frac{dg(n)}{dn} = g(n) h(\log n),
\end{align}
we obtain
\begin{align}
  \frac{d\log g(n)}{d\log n} = h(\log n)
\end{align}
and consequently
\begin{align}
  \label{HUGN}
  g(n)=\exp\okra{\int_0^{\log n}h(u)du}.
\end{align}
In particular, the maximal number of types is finite if
\begin{align}
\int_0^\infty h(u)du<\infty.
\end{align}
Hence a quick glance at the plot of function $h(u)$ with respect to
variable $u=\log n$ can inform a guess whether the potential
vocabulary of a given text is finite or not. For example, if function
$h(u)$ follows a decaying linear trend then we may guess that the
potential vocabulary is finite.

In consequence, to derive the smoothed vocabulary size $g(n)$ and the
smoothed rank function $g(n||f)$, it suffices to assume a certain form
of the smoothed hapax rate $h(u)$. Since function $h(\log n)$ varies
slowly as a function of the text length $n$, it seems a convenient
object for approximations, as we will discuss thoroughly in \S
\ref{secModels}.  A necessary, though not sufficient, requirement for
the smoothed hapax rate is that it remains in the unit interval,
$0\le h(u)\le 1$.  For example, for a constant function
$h(u)=\beta\in(0,1)$, we obtain a power-law growth of the vocabulary
$g(n)=\exp(\beta\log n)=n^\beta$, known as Herdan-Heaps' law
\citep{Herdan64,Heaps78} mentioned in \S \ref{secPerspective}. This
model is a sort of a baseline to be analyzed in \S \ref{secConstant}.

Before we delve into particular examples of the smoothed hapax rate,
let us comment on the necessary conditions for this object.  As a
consequence of formulae (\ref{Derivatives}) and (\ref{HUGN}), we can
express the smoothed frequency spectrum and the smoothed rank function
as
\begin{align}
  g(n|k)&= g(n) h(\log n|k),
  \\
  g(n||f)&= -g(n) \sum_{k=0}^{f-1} h(\log n|k),
\end{align}
where we define recursively $h(u|0):=-1$ and
\begin{align}
  h(u|k)
  &:=
    \kwad{1-\frac{1}{k}\okra{1+h(u)+\frac{d}{du}}}h(u|k-1),
    \quad k\ge 1.
    \label{Recursion}
\end{align}
We notice that $h(u|1)=h(u)$.  Recursion (\ref{Recursion}) was
observed by \citet[equation (3.43)]{Baayen01}. Functions $h(u|k)$ are
called the \emph{relative spectrum elements} \citep[page
90]{Baayen01}.

The necessary conditions (\ref{GConditions}) are equivalent to
conditions
\begin{align}
  \int_{-\infty}^0 h(u)du&=\infty,
  &
  h(u|k)&\ge 0 \text{ for } k\ge 1.
  \label{HConditions}
\end{align}
We note that if function $u\mapsto h(u)$ satisfies conditions
(\ref{HConditions}) then so does function $u\mapsto h(u-\alpha)$ but
we do not know whether function $u\mapsto h(\gamma u)$ can be an
admissible smoothed hapax rate.  This is bad news since we do not have
an easy theoretical control of the slope of the smoothed hapax rate
function.

We note in passing that the necessary conditions (\ref{HConditions})
for the smoothed hapax rate $h(u)$ are unrelated to the log-log
concavity of the vocabulary size, empirically observed by
\citet{FontClosCorral15}. The log-log concavity of the smoothed
vocabulary size $g(n)$ is equivalent to the monotonous decrease of the
smoothed hapax rate, which can be stated as
\begin{align}
  \frac{dh(u)}{du}=\frac{d^2\log g(n)}{d(\log n)^2}<0.
\end{align}
This holds since a function is concave if and only if its second
derivative is negative. Mutatis mutandis, the analogous statement is
valid for the log-log concavity.  By contrast, as noticed by
\citet{Fan10}, the empirical hapax rate is $U$-shaped for large
corpora, which can be easily modeled by mixture models to be described
in \S \ref{secMixture}.

\section{Hapax rate models}
\label{secModels}

In this section, a model will be understood as a particular choice of
smoothed hapax rate $h(u)$ and implied functions $g(n)$, $g(n|k)$, and
$g(n||f)$ that follow by the formulae derived in \S \ref{secAnalytic}.
The empirical hapax rate $v_1/v$ for texts of the size of novels, to
be discussed in \S \ref{secExperiments}, has usually a decaying
shape. It equals $1$ for the text length $n=1$ and it decays slowly
for the growing argument.  The dominating trend is approximately
linear in terms of variable $u=\log n$ and the slope of the empirical
hapax rate is of an approximate magnitude $-0.05$.  By contrast, for
much larger corpora studied by \citet{Fan10}, the empirical hapax rate
is $U$-shaped.

Having this in mind, we will consider four models of the smoothed
hapax rate that apply to novel-sized texts:
\begin{align}
  \label{DConstant}
  h_\beta(u)&=\beta,
  &&\text{(constant model)}
  \\
  \label{DCancelation}
  h_\alpha(u)&=\frac{1}{u-\alpha}-\frac{1}{e^{u-\alpha}-1},
  &&\text{(cancelation model)}
  \\
  \label{DLinear}
  h_{\alpha\gamma}(u)
            &=
              \begin{cases}
                1, & u<\alpha,
                \\
                1-\gamma (u-\alpha), & \alpha\le u\le \gamma^{-1}+\alpha,
                \\
                0, & u>\gamma^{-1}+\alpha,
              \end{cases}
  &&\text{(linear model)}
  \\
  \label{DLogistic}
  h_{\alpha\beta\gamma}(u)&=\frac{1-\beta}{1+e^{\gamma
                            (u-\alpha)}}+\beta.
  &&\text{(logistic model)}
\end{align}

We will show that the constant model is a crude baseline that implies
the exact Herdan-Heaps law for any text size. The cancelation model
implies the exact Zipf law for a certain text size and reproduces the
general decreasing trend of the hapax rate. The linear model is a more
precise model of the hapax rate than the cancelation model but it is
not analytic. The logistic model corrects on this issue and
additionally it allows to model the saturation of the hapax rate for
increasing text sizes, which may be necessary for large corpora.  Some
of these models were partly discussed in literature.  In \S
\ref{secConstant}--\S \ref{secLogistic}, we will derive corresponding
functions $g(n)$, $g(n|k)$, and $g(n||f)$ for these four models,
giving more detailed references to the preceding mentions of derived
formulae.

\subsection{Constant model}
\label{secConstant}

The constant model consists in a constant smoothed hapax rate,
\begin{align}
  \label{HerdanH}
  h(u)=\beta\in(0,1].
\end{align}
This model is a baseline since the argument of the smoothed hapax rate
equals variable $u=\log n$, which varies slowly with the text size
$n$. Thus, in some relatively large region, we may assume that $h(u)$
is approximately constant.  A special case of the constant model is
the maximal model for $\beta=1$, where
\begin{align}
  h(u)&:=1,
  &
  g(n)&:=n,
  \\
  g(n|k)&:=\boole{k=1}n,
  &
  g(n||f)&:=\boole{f=1}n.
\end{align}
The maximal model arises when all tokens are of distinct types.

Now let us solve the constant model for $\beta\neq 1$.  By formula
(\ref{HUGN}), the constant model (\ref{HerdanH}) implies a power-law
growth of the smoothed vocabulary size,
\begin{align}
  \label{HerdanG}
  g(n)=\exp(\beta\log n)=n^\beta.
\end{align}
As mentioned in \S \ref{secPerspective}, this is known as
Herdan-Heaps' law \citep{Herdan64,Heaps78}.  In view of the one-to-one
correspondence between functions $h(u)$ and $g(n)$, Herdan-Heaps' law
is equivalent to a constant rate of hapaxes, regardless of the text
size.

Let us define the binomial coefficients with a real upper argument,
\begin{align}
  \binom{r}{k}:=\frac{r(r-1)...(r-k+1)}{k!}.
\end{align}
Differentiating function $g(n)=n^\beta$, we obtain the frequency
spectrum
\begin{align}
  g(n|k)=(-1)^{k+1}\binom{\beta}{k}\,n^\beta
  =-\binom{k-\beta-1}{k}\,n^\beta.
\end{align}

Applying the Taylor expansion (\ref{Taylor}), by induction, we derive
the smoothed rank function
\begin{align}
  \label{MandelbrotCorrected}
  g(n||f)
  =
  g(n)-\sum_{k=1}^{f-1} g(n|k)
  =
  \binom{f-\beta-1}{f-1}\,n^\beta.
\end{align}
We observe that for the constant model, the normalized smoothed rank
function $g(n||f)/g(n)$ is invariant with respect to the text size! It
is a sort of a scale-free distribution.  By a scale-free distribution,
we mean a distribution which does not distinguish any particular text
length or a vocabulary size.

We may approximate the smoothed rank function
(\ref{MandelbrotCorrected}) as
\begin{align}
  \log g(n||f)
  % &=
  % \log \binom{f-\beta-1}{f-1}+\beta\log n
  %   \nonumber\\
  &=
  \sum_{k=1}^{f-1} \log\okra{1-\frac{\beta}{k}}+\beta\log n
  \nonumber\\
  &\approx
    -\sum_{k=1}^{f-1} \frac{\beta}{k}+\beta\log n
    \approx
  \beta\okra{\log \frac{n}{f-1}-\gamma}.
  \label{MandelbrotAsymptotic}
\end{align}
Approximation (\ref{MandelbrotAsymptotic}) of formula
(\ref{MandelbrotCorrected}) resembles Mandelbrot's formula
(\ref{Mandelbrot}) in the middle-rank region.  Out of the middle-rank
region, the smoothed rank function (\ref{MandelbrotCorrected}) is
somewhat different than Mandelbrot's heuristic correction
(\ref{Mandelbrot}).  The discrepancy between formulae
(\ref{Mandelbrot}) and (\ref{MandelbrotCorrected}) arises both for
very large and very small ranks. In particular, the number of hapaxes
is significantly larger for the constant model than for Mandelbrot's
formula (\ref{MandelbrotCorrected}). This was already noticed by
\citet{Baayen01}.

What kind of a memoryless source generates texts that follow the
constant model? Inverting approximation (\ref{MandelbrotAsymptotic}),
we obtain
\begin{align}
  \log\frac{f_r-1}{n}\approx -\frac{1}{\beta}\log r-\gamma.
\end{align}
Using the theory developed in \S \ref{secConvergence}, we see that the
theoretical probabilities satisfy $p_r=\lim_{n\to\infty} f_r/n$. Hence
we obtain the asymptotic power-law probability distribution
\begin{align}
  \lim_{r\to\infty} \frac{p_r}{r^{-1/\beta}e^{-\gamma}}=1,
\end{align}
where $\gamma$ is the previously discussed Euler gamma.

\subsection{Cancelation model}
\label{secDavis}

In \S \ref{secConstant}, we saw an example of a scale-free word
frequency distribution.  In this section, we will develop a model
described by \citet[pages 97--101]{Baayen01} and rediscovered by
\citet{Davis18}, which is not scale invariant and which reproduces the
exact Zipf law for the text size $n=1$. (Zipf's law for text size
$n=1$ should be understood as a mathematical idealization in the
framework of analytic functions, where the text length and the
spectrum elements are real numbers and the unit of the text length can
be chosen arbitrarily.) We call this model the cancelation model to
have a more neutral and distinctive name. \citet[Chapter 3]{Baayen01}
considered many more related models based on series
(\ref{GN})--(\ref{GNk}) with a particular choice of the frequency
spectrum for an ideal text size $n$. The cancelation model is a
special case of the Yule-Simon model \citep[pages 107-114]{Baayen01},
which in turn specializes the Waring-Herdan-Muller model \citep[pages
114-117]{Baayen01}.

The cancelation model starts with an apparently eccentric formula for
the smoothed hapax rate,
\begin{align}
  \label{DavisH}
  h(u)=\frac{1}{u}-\frac{1}{e^{u}-1},
\end{align}
where $\lim_{u\to -\infty} h(u)=1$, $\lim_{u\to 0} h(u)=1/2$, and
$\lim_{u\to \infty} h(u)=0$. Thus function (\ref{DavisH}) has a
decaying sigmoid shape. In particular, we chose name ``cancelation
model'' for function (\ref{DavisH}) since both functions
$u\mapsto 1/u$ and $u\mapsto 1/(e^{u}-1)$ have poles at $u=0$ that get
canceled.  To fit the cancelation model to real data, we have to
rescale it using the offset operation
\begin{align}
  \label{OffsetOne}
  h_\alpha(u)&:= h(u-\alpha),
  &
    g_\alpha(n)&:=\frac{g(ne^{-\alpha})}{g(e^{-\alpha})},
  \\
  \label{OffsetTwo}
  g_\alpha(n|k)&:=\frac{g(ne^{-\alpha}|k)}{g(e^{-\alpha})},
    &
      g_\alpha(n||f)&:=\frac{g(ne^{-\alpha}||f)}{g(e^{-\alpha})}.
\end{align}
where $\alpha$ is an empirically chosen real parameter.  Similar
offset operation with parameter $\alpha$ will be applied tacitly to
all subsequently discussed models. Obviously, parameter $\alpha$ makes
sense only for word frequency distributions that are not scale-free,
i.e., when $h(u)$ is not constant. Exactly, $\alpha$ is a location
parameter that selects a certain particular text length.

By (\ref{HUGN}), the cancelation model (\ref{DavisH}) implies an
asymptotically logarithmic vocabulary growth,
\begin{align}
  \label{DavisG}
  g(n)
  =\frac{n\log n}{n-1}\approx \log n. 
\end{align}
To deal with the smoothed vocabulary size $g(n)$, let us consider the
series
\begin{align}
  \sum_{m=1}^\infty \frac{x^m}{m(m+1)}
  =1+\frac{(1-x)\log(1-x)}{x},
  \quad
    \abs{x}\le 1.
\end{align}
Hence we may write the smoothed vocabulary size and the smoothed
frequency spectrum as
\begin{align}
  g(n)&=1-\sum_{m=1}^\infty \frac{(1-n)^m}{m(m+1)},
  \\
  g(n|k)&=
          \sum_{m=k}^\infty \binom{m}{k} \frac{n^{k}(1-n)^{m-k}}{m(m+1)},
          \quad
          k\ge 1.
\end{align}
Thus for the cancelation model (\ref{DavisH}) and text size $n=1$
(which should be understood purely formally and always rescaled via
the offset operation (\ref{OffsetOne})--(\ref{OffsetTwo}) in any
meaningful application), we obtain Lotka's law (\ref{Lotka}) and
Zipf's law (\ref{Zipf}) in a rescaled form,
\begin{align}
  g(1|k)&=\frac{1}{k(k+1)},
  \\
    g(1||f)&=\frac{1}{f}. 
\end{align}

Now we will evaluate the smoothed rank function $g(n||f)$ for any text
size $n$.  \citet{Davis18} evaluated the derivatives of $g(n)$ using a
computer algebra system without noticing a simple pattern that emerges
from the calculations. Let us present a brief derivation of this
regularity.  We can write the derivatives
\begin{align}
  \label{GPQ}
  g^{(k)}(n)
  &=\sum_{j=0}^k \binom{k}{j} p^{(j)}(n) q^{(k-j)}(n)
    ,
\end{align}
where
\begin{align}
  p^{(0)}(n)&=n\log n,
  & 
  p^{(1)}(n)&=1+\log n,
  \\
  p^{(j)}(n)&=\frac{(-1)^{j}(j-2)!}{n^{j-1}},
                   \quad
                   j\ge 2,
  &
  q^{(j)}(n)&=\frac{(-1)^{j}j!}{(n-1)^{j+1}}.
\end{align}

Let us define $s_{-1}:=0$, $s_{0}:=\log n$, $s_{j}:=-1/j$ for
$j\ge 1$, and $t:=1-1/n$.  Hence we may compute the smoothed
rank function as
\begin{align}
  g(n||f)
  &=\sum_{k=0}^{f-1}\sum_{j=0}^k
    \frac{(-1)^jn^{j-1}p^{(j)}(n)}{j!}
    \cdot
    \frac{(-1)^{k-j}n^{k-j+1}q^{(k-j)}(n)}{(k-j)!}
    \nonumber\\
  &=\sum_{k=0}^{f-1}\sum_{j=0}^k \kwad{s_{j}-s_{j-1}} t^{j-k-1}
    %\nonumber\\
  %&
    =\sum_{j,i\ge 0:\,j+i\le f-1} \kwad{s_{j}-s_{j-1}} t^{-i-1}
    \nonumber\\
  &=\sum_{j,i\ge 0:\,j+i=f-1} s_{j} t^{-i-1}
  =\sum_{j=0}^{f-1} s_{j} t^{j-f}
  .
    \label{ZipfCorrectedFast}
\end{align}
Subsequently, we may rewrite formula (\ref{ZipfCorrectedFast}) as
expression
\begin{align}
  g(n||f)
  &=\frac{\log n-\sum_{j=1}^{f-1} \okra{1-1/n}^{j}/j
    }{\okra{1-1/n}^{f}}
    .
   \label{ZipfCorrected}
\end{align}
Formula (\ref{ZipfCorrected}) adjusts the ideal Zipf law for an
arbitrary text size.

As for the practical use of formula (\ref{ZipfCorrected}), we notice
the Taylor series
\begin{align}
  \log x = \sum_{j=1}^\infty \frac{\okra{1-1/x}^{j}}{j},
  \quad
  x\ge \frac{1}{2},
\end{align}
so the limit of expression (\ref{ZipfCorrected}) for $n\to 1$ is
indeed $1/f$, i.e., the ideal Zipf law. If we could compute the
logarithm function with an arbitrarily high precision then formula
(\ref{ZipfCorrected}) would imply an efficient algorithm for computing
the smoothed rank function plot for an arbitrary text length $n$. It
only takes $O(l)$ arithmetic operations to compute smoothed ranks
$g(n||f)$ for all frequencies $f\le l$. Indeed, for $n\neq 1$, we may
write the recursion
\begin{align}
  \label{ZipfRecursion}
  g(n||1)&= \frac{\log n}{1-1/n},
  \\
  \label{ZipfRecursionII}
  g(n||f+1)&= \frac{g(n||f)-1/f}{1-1/n},
       \quad
       f\ge 1,
\end{align}
whereas for $n\ge 1/2$, we may also consider summing the series
\begin{align}
  g(n||f)=\sum_{j=0}^\infty \frac{\okra{1-1/n}^{j}}{j+f},
  \label{ZipfCorrectedSeries}
\end{align}
which is considerably slower but more precise. We notice some beauty
of recursion (\ref{ZipfRecursionII}), where numerator $g(n||f)-1/f$ is
the deviation of the smoothed rank function from the exact
Zipf law and denominator $1-1/n$ is the deviation of the text size
from the ideal text size.

What kind of a memoryless source generates texts that follow the
cancelation model? Formula (\ref{ZipfCorrectedSeries}) for large $n$
can be approximated as
\begin{align}
  g(n||f)
  &\approx
  \sum_{j=0}^\infty
    \frac{\exp\okra{-{j}/{n}}}{{j}/{n}+{f}/{n}}\cdot\frac{1}{n}
  \nonumber\\
  &\approx
  \int_0^\infty \frac{\exp(-u)}{u+{f}/{n}}du
  =
  \exp\okra{\frac{f}{n}}\Gamma\okra{0,\frac{f}{n}},
\end{align}
where the incomplete gamma function is
\begin{align}
  \Gamma(a,x):=\int_x^\infty t^{a-1}\exp(-t)dt.
\end{align}
Hence, for the theoretical probabilities
$p_r=\lim_{n\to\infty} f_r/n$, we obtain
\begin{align}
  \exp\okra{p_r}\Gamma\okra{0,p_r}\approx r.
\end{align}
Since we have 
\begin{align}
  \Gamma(0,x)=-\gamma-\log x-\sum_{j=0}^\infty\frac{(-x)^j}{j(j!)},
\end{align}
then for small probabilities $p_r$, we obtain
$\exp\okra{p_r}\Gamma\okra{0,p_r}\approx -\gamma-\log p_r$. Hence the
theoretical probabilities decay exponentially asymptotically, 
\begin{align}
  \lim_{r\to\infty} \frac{p_r}{e^{-r-\gamma}}=1,
\end{align}
where $\gamma$ is the previously discussed Euler gamma. It should be
stressed that this approximation holds only in the limit, whereas for
small ranks, we probably have a behavior closer to Zipf's law
$p_r\approx C/r$.

\subsection{Linear model}
\label{secLinear}

To keep the theory consistent, functions $g(n)$ and $h(u)$ are assumed
to be analytic functions that satisfy conditions (\ref{GConditions})
and (\ref{HConditions}), respectively.  On the other hand, as we have
mentioned, the empirical hapax rate has usually a decaying shape for
moderately sized texts. Moreover, as the empirical data to be
discussed in \S \ref{secExperiments} indicate, this trend is
approximately linear in terms of variable $u=\log n$.  Therefore, we
may be tempted to propose the following piecewise linear model of the
smoothed hapax rate,
\begin{align}
  \label{LinearH}
  h(u)&=
        \begin{cases}
          1, & u<0,
          \\
          1-\gamma u, & 0\le u\le \gamma^{-1},
          \\
          0, & u>\gamma^{-1}.
        \end{cases}
\end{align}
where $\gamma>0$. Of course, this model is not an analytic function
and we have no guarantee that conditions (\ref{GConditions}) and
(\ref{HConditions}) are satisfied even for $0\le u\le \gamma^{-1}$.

Nonetheless, let us proceed with calculations and let us derive the
corresponding ill-defined smoothed vocabulary size $g(n)$ and the
smoothed relative spectrum elements $h(u|k)$. By (\ref{HUGN}), we have
\begin{align}
  g(n)&=
        \begin{cases}
          n, & n\le 1,
          \\
          n^{1-\frac{1}{2}\gamma\log n}, & 1\le n\le \exp(\gamma^{-1}),
          \\
          \sqrt{\exp(\gamma^{-1})}, & n>\exp(\gamma^{-1}).
  \end{cases}
\end{align}
Thus the smoothed vocabulary size is bounded by a finite constant. For
the empirically motivated value of slope $\gamma\approx 0.05$, the
maximal vocabulary size is $\sqrt{\exp(\gamma^{-1})}\approx 22\,026$
types and the text length for which this limit is hit amounts to
$\exp(\gamma^{-1})\approx 4.85\cdot 10^8$ tokens. We note that this
maximal vocabulary size is close in the magnitude to the ideal size of
the vocabulary estimated in (\ref{TypesZipf}).

For $0\le u\le \gamma^{-1}$, recursion (\ref{Recursion}) implies that
the relative spectrum elements are polynomials,
\begin{align}
  \label{LinearPoly}
  h(u|k)&=\frac{1}{k!} \sum_{m=0}^k a_{km}u^m,
\end{align}
where we have the recursion $a_{km}=0$ for $m<0$ or $m>k$,
$a_{00}=-1$, and
\begin{align}
  \label{LinearRecursion}
  a_{km}&=
          \gamma a_{k-1,m-1}+(k-2)a_{k-1,m}-(m+1)a_{k-1,m+1}
\end{align}
for $k\ge 1$ and $0\le m\le k$.

From (\ref{LinearPoly}), we can compute the smoothed rank function as
explained in \S \ref{secAnalytic}.  We have not seen a clear pattern
in these polynomials that would solve the recursion in a closed form.
We do not know what the range of $u$ is such that $h(u|k)\ge 0$ for
all $k\ge 1$.

\subsection{Logistic model}
\label{secLogistic}

The plain logistic model is unrealistic as a model of word frequency
distributions because of a too large slope but it can be analyzed
easily and, after a modification, it yields the best model that we
propose in this paper.  The plain logistic model assumes the familiar
logistic function as the smoothed hapax rate,
\begin{align}
  \label{LogisticH}
  h(u)=\frac{1}{1+e^{u}},
\end{align}
where $\lim_{u\to -\infty} h(u)=1$, $\lim_{u\to 0} h(u)=1/2$, and
$\lim_{u\to \infty} h(u)=0$. Hence function (\ref{LogisticH}) has also
a decaying sigmoid shape.

By (\ref{HUGN}), the logistic model (\ref{LogisticH}) implies a
bounded smoothed vocabulary size,
\begin{align}
  \label{LogisticG}
  g(n)
  =\frac{2n}{n+1}\xrightarrow[n\to\infty]{} 2.
\end{align}
Let us evaluate the corresponding smoothed rank function.  Like for
the cancelation model, we can write the derivatives as (\ref{GPQ})
where
\begin{align}
  p^{(0)}(n)&=2n,
  &
  p^{(1)}(n)&=2,
  \\
  p^{(j)}(n)&=0,
                   \quad
                   j\ge 2,
  & 
    q^{(j)}(n)&=\frac{(-1)^{j}j!}{(n+1)^{j+1}}.
\end{align}
Hence we may compute the smoothed rank function as
\begin{align}
  g(n||f)
  &=g(n)+\sum_{k=1}^{f-1}\sum_{j=0}^1
    \frac{(-1)^jn^{j-1}p^{(j)}(n)}{j!}
    \cdot
    \frac{(-1)^{k-j}n^{k-j+1}q^{(k-j)}(n)}{(k-j)!}
  \nonumber\\
  &=\frac{2n}{n+1}+
    \sum_{k=1}^{f-1}\kwad{\frac{2n^{k+1}}{(n+1)^{k+1}}
    -\frac{2n^{k}}{(n+1)^{k}}}
  %\nonumber\\
  =\frac{2n^{f}}{(n+1)^{f}}.
\end{align}
Thus the rank function is a geometric progression. Conditions
(\ref{GConditions}) and (\ref{HConditions}) are satisfied.

The linear model discussed in \S \ref{secLinear} suggests a
certain amendment of the logistic model, which makes it more
realistic. This modification is analytic in contrast to the linear
model. For analytic functions $h(u)$ and $g(n)$, in general, we may
try parameterizing
\begin{align}
  \label{ScalingH}
  h_{\gamma\beta}(u)&:=(1-\beta)h(\gamma u)+\beta,
  \\
  \label{ScalingG}
  g_{\gamma\beta}(n)&:= [g(n^\gamma)]^{(1-\beta)/\gamma}n^\beta,
\end{align}  
with parameters $\gamma>0$ and $0\le\beta<1$ but it is not guaranteed
that the smoothed frequency spectrum elements still satisfy
$h_{\gamma\beta}(u|k)\ge 0$ and $g_{\gamma\beta}(u|k)\ge 0$. Anyway,
since we are in need to control the slope of the smoothed hapax rate
$h(u)$ and we wish it to be approximately linear, let us apply scaling
operation (\ref{ScalingH})--(\ref{ScalingG}) to (\ref{LogisticH}) and
try the rescaled sigmoid function
\begin{align}
  \label{SLogisticH}
  h_{\gamma\beta}(u)=\frac{1-\beta}{1+e^{\gamma u}}+\beta.
\end{align}

By (\ref{HUGN}), the rescaled model (\ref{SLogisticH}) implies an
asymptotically power-law growing smoothed vocabulary size,
\begin{align}
  \label{SLogisticG}
  g_{\gamma\beta}(n)
  =\frac{2^{(1-\beta)/\gamma}n}{(n^\gamma+1)^{(1-\beta)/\gamma}}
  \approx 2^{(1-\beta)/\gamma} n^\beta.
\end{align}
Taking derivatives of function (\ref{SLogisticG}) can be a tedious
exercise. Like for the linear model, it is simpler to deal with
recursion (\ref{Recursion}) for the relative spectrum elements.
Cleverly, let us observe that the first derivative of function
(\ref{SLogisticH}) for $\beta=0$ is a polynomial of $h_{\gamma 0}(u)$,
\begin{align}
  \label{SLogisticHD}
  \frac{d}{du} h_{\gamma 0}(u)=-\gamma
  h_{\gamma 0}(u)[1-h_{\gamma 0}(u)].
\end{align}
% The value of the first derivative for $u=0$ and $\beta=0$ is
% $-\gamma/4$. It is the minimum.  Since the slope of $h(u)$ for
% natural language is close to $-0.05$ then the optimal $\gamma$
% should equal $0.2$ approximately.  Table \ref{tabPars} informs us
% that the optimal $\gamma$ is close to $0.3$, however.

In view of observation (\ref{SLogisticHD}) and recursion
(\ref{Recursion}), rewritten as
\begin{align}
  h_{\gamma\beta}(u|k)
  &:=
    \kwad{1-\frac{1}{k}\okra{1+\beta+(1-\beta)h_{\gamma 0}(u)+
    \frac{d}{du}}}h_{\gamma\beta}(u|k-1) 
\end{align}
for $k\ge 1$, the relative spectrum elements $h_{\gamma\beta}(u|k)$
are polynomials of variable $h_{\gamma 0}(u)$,
\begin{align}
  \label{SLogisticPoly}
  h_{\gamma\beta}(u|k)&=\frac{1}{k!} \sum_{m=0}^k b_{km}[h_{\gamma 0}(u)]^m.
\end{align}
Since we can compute the derivative of a power of $h_{\gamma 0}(u)$ as
\begin{align}
  \frac{d}{du} [h_{\gamma 0}(u)]^m
  &=m [h_{\gamma 0}(u)]^{m-1} \frac{d}{du} h_{\gamma 0}(u)
  %\nonumber\\
  =-\gamma m [h_{\gamma 0}(u)]^m[1-h_{\gamma 0}(u)],
\end{align}
we obtain the recursion $b_{km}=0$ for $m<0$ or $m>k$, $b_{00}=-1$, and 
\begin{align}
  \label{SLogisticRecursion}
  b_{km}&=
          (-1+\beta-\gamma (m-1)) b_{k-1,m-1}+(k-1-\beta+\gamma m)b_{k-1,m}
\end{align}
for $k\ge 1$ and $0\le m\le k$.  From (\ref{SLogisticPoly}), we can
compute the smoothed rank function as explained in \S
\ref{secAnalytic}.

\section{Experiments}
\label{secExperiments}

Consequently, we proceed to an experimental verification of our models
by comparing them with empirical word frequency distributions. There
is a formidable amount of textual data available in various languages
on the internet. Aware of this, we restrict ourselves to a pilot test
of our models. We are not expecting to confirm the absolute truth of
any particular model. We confine to setting up an experimental
methodology that can be extended in future research. We are
temporarily satisfied with a general conclusion that the urn model
predicts word frequency distributions reasonably well if we plug in a
simple parametric hapax rate model. Our goal is to explore a few such
hapax rate models and to propose a way of looking at word frequency
distributions that immediately suggests how to correct the ideal Zipf
law and some of its later corrections. We observe that one can easily
improve considerably over the constant hapax rate
model. Simultaneously, by choosing a monotonously decreasing hapax
rate model, one can model the empirical log-log concavity of the
vocabulary size plot, which the constant hapax rate model cannot
capture. This can be all achieved under the standard urn model. We
suppose that a similar framework can be applied to large corpora in
the future, where the shape of the hapax rate is $U$-shaped.

\subsection{Data}
\label{secData}

As our data, we considered 14 texts in English downloaded from Project
Gutenberg (\url{https://www.gutenberg.org/}). The texts are listed in
Table \ref{tabTexts}. We list the names of the corresponding text
files since, for brevity, we use them in the following tables to
identify the texts. We did not perform a deep preprocessing of the
texts. To somewhat normalize the texts, the Gutenberg Project
preambles were removed and the characters in the text files were
projected onto 26 capital letters and a space---27 distinct characters
in total as in \citep{Shannon51}. The sequences of punctuation
characters and spaces were replaced by a single space and all other
non-letter characters were replaced by letter X.  No lemmatization was
done and word tokens were identified with strings delimited by
spaces. The resulted texts ranged in the length between 90\,000 and
1\,300\,000 word tokens.

\subsection{Methods}
\label{secMethods}

In the following, we describe our method of fitting smoothed hapax
rate models to empirical data. Broadly speaking, we tried first to
construct an empirically-driven object that can be reasonably compared
with a smoothed hapax rate model.

For each considered text
  $\mathbf{t}^*=(t^*_1,t^*_2,...,t^*_{n^*})$, where $t^*_i$ is the
  $i$-th word token, we began with computing the table of frequencies
  for all word types,
\begin{align}
  f^*(w):=\sum_{i=1}^{n^*} \boole{t^*_i=w}.
\end{align}
Given that, we computed the rank function
\begin{align}
  r^*_k:=\sum_w\boole{f^*(w)\ge k},
\end{align}
the frequency spectrum elements $v^*_k:=r^*_k-r^*_{k+1}$, the number
of tokens $n^*:=\sum_{k=1}^{\infty} k v^*_k$, and the number of types
$v^*:=\sum_{k=1}^{\infty} v^*_k$.

Consequently, we computed the smoothed number of types $g^*(n)$
  and the smoothed number of hapaxes $g^*(n|1)$ defined via
\begin{align}
  \label{GNV}
  g^*(n)&:=v^*-\sum_{k^*=1}^{n^*} v^*_{k^*} \okra{1-\frac{n}{n^*}}^{k^*},
  \\
  \label{GNVk}
  g^*(n|1)&:=\sum_{k^*=1}^{n^*} k^* v^*_{k^*}
          \okra{\frac{n}{n^*}}\okra{1-\frac{n}{n^*}}^{k^*-1}.
\end{align}
The corresponding smoothed hapax rate is
$h^*(u):=g^*(e^u|1)/g^*(e^u)$. These formulae specialize
(\ref{GN})--(\ref{GNk}) to the case of $g(n^*)=v^*$ and
$g(n^*|1)=v^*_1$.

The smoothed number of types $g^*(n)$ and the smoothed number of
hapaxes $g^*(n|1)$ should be distinguished from the incremental number
of types $g^\#(n)$ and the incremental number of hapaxes $g^\#(n|1)$,
which were also computed for comparison. These functions are defined
as $g^\#(n):=\card W(n)$ and $g^\#(n|1):=\card W(n|1)$, where $W(n)$
is the set of types in the first $n$ tokens of the text and $W(n|1)$
is the set of hapaxes in the first $n$ tokens of the text.  The
corresponding incremental hapax rate is
$h^\#(u):=g^\#(e^u|1)/g^\#(e^u)$.

Whereas functions $g^\#(n)$ and $g^\#(n|1)$ fluctuate strongly,
functions $g^*(n)$ and $g^*(n|1)$ are smooth. We decided to fit models
defined in \S \ref{secModels} to the plot of the smoothed number of
types $g^*(n)$ defined in (\ref{GNV}). We have observed that fitting
the models to the smoothed function $g^*(n)$ yields a visually better
prediction of the empirical rank function $r^*_f$ than fitting them to
the incremental number of types $g^\#(n)$.  Similarly, fitting the
smoothed number of types $g^*(n)$ yielded a visually better prediction
of the empirical rank function $r^*_f$ than fitting the smoothed hapax
rate $g^*(n|1)/g^*(n)$. In the following, we present the results of
fitting the models from \S \ref{secConstant}--\S \ref{secLogistic}
only to the smoothed number of types $g^*(n)$.

\subsection{Results}
\label{secResults}

\begin{table}[t]
\caption{The selection of texts from Project
  Gutenberg.\label{tabTexts}}
\centering
\medskip
\begin{tabular}{|l|l|r|}
  \toprule
  Title &Author &File \\
  \midrule
  First Folio/35 Plays &W. Shakespeare & {00ws110.txt} \\
  One of Ours&W. Cather & {1ours10.txt} \\
  20,000 Leagues under the &J. Verne & {2000010.txt} \\
  \hfill Sea & & \\
  Critical \& Historical Essays &Macaulay & {2cahe10.txt} \\
  Five Weeks in a Balloon &J. Verne & {5wiab10.txt} \\
  Eight Hundred Leagues on &J. Verne & {800lg10.txt} \\
  \hfill the Amazon & & \\
  The Complete Memoirs &J. Casanova & {csnva10.txt} \\
  Memoirs &Comtesse du Barry & {dbrry10.txt} \\
  The Descent of Man &C. Darwin & {dscmn10.txt} \\
  Gulliver's Travels &J. Swift & {gltrv10.txt} \\
  The Mysterious Island &J. Verne & {milnd10.txt} \\
  Mark Twain, A Biography &A. B. Paine & {mt7bg10.txt} \\
  The Journal to Stella &J. Swift & {stlla10.txt} \\
  Life of William Carey &G. Smith & {wmcry10.txt} \\
  \botrule
\end{tabular}  
\end{table}

\begin{table*}[t]
  \caption{The parameters fitted by least squares to function
    $g^*(n)$.\label{tabPars}}
  \centering
  \medskip
  \footnotesize
  \begin{tabular}{|l|l|l|lll|ll|r|}
    \toprule
    File & Constant & Cancelation
    & \multicolumn{3}{c|}{Logistic}
    & \multicolumn{2}{c|}{Linear} & Length\\
    \cline{2-2}\cline{3-3}\cline{4-6}\cline{7-8}
    & $\beta$       & $\alpha$
    & $\gamma$      & $\beta$       & $\alpha$
    & $\gamma$      & $\alpha$      & $n^*$\\
\midrule
00ws110.txt     & 0.768 & 12.06 & 0.314 & 0.218 & 10.11 & 0.0509        & 2.14  & 835726\\
1ours10.txt     & 0.797 & 11.55 & 0.318 & 0.203 & 9.72  & 0.0507        & 1.7   & 128963\\
2000010.txt     & 0.801 & 11.48 & 0.323 & 0.008 & 10.62 & 0.0578        & 2.22  & 101247\\
2cahe10.txt     & 0.796 & 12.12 & 0.314 & 0     & 11.38 & 0.0576        & 2.79  & 298339\\
5wiab10.txt     & 0.808 & 11.64 & 0.315 & 0.001 & 10.86 & 0.0552        & 2.13  & 92558\\
800lg10.txt     & 0.799 & 11.43 & 0.327 & 0.162 & 9.77  & 0.0534        & 1.84  & 95493\\
csnva10.txt     & 0.732 & 11.39 & 0.308 & 0.157 & 9.94  & 0.0542        & 1.87  & 1268149\\
dbrry10.txt     & 0.787 & 11.39 & 0.325 & 0.065 & 10.31 & 0.0583        & 2.23  & 159710\\
dscmn10.txt     & 0.774 & 11.5  & 0.328 & 0     & 10.75 & 0.0629        & 2.71  & 312075\\
gltrv10.txt     & 0.796 & 11.4  & 0.322 & 0.001 & 10.62 & 0.0584        & 2.22  & 104909\\
milnd10.txt     & 0.773 & 11.14 & 0.347 & 0.127 & 9.63  & 0.0608        & 2.24  & 195064\\
mt7bg10.txt     & 0.775 & 11.91 & 0.296 & 0.001 & 11.45 & 0.0565        & 2.55  & 519886\\
stlla10.txt     & 0.757 & 10.91 & 0.333 & 0.231 & 8.87  & 0.0536        & 1.45  & 245882\\
wmcry10.txt     & 0.799 & 11.69 & 0.314 & 0     & 10.96 & 0.0567        & 2.34  & 145487\\
\midrule
Mean    & 0.783 & 11.54 & 0.32  & 0.084 & 10.36 & 0.0562        & 2.17  & 321678\\
\botrule
  \end{tabular}
\end{table*}

\begin{table}[t]
  \caption{The goodness of fit $\sqrt{\text{WSSR/ndf}}$ for
    function $g^*(n)$.\label{tabWSSRs}}
  \centering
  \medskip
  \begin{tabular}{|l|r|r|r|r|}
    \toprule
    File    & Constant      & Cancelation & Logistic      & Linear\\
    \midrule
    00ws110.txt     & 1784.34       & 120.42        & \textbf{11.82}        & 43.71\\
    1ours10.txt     & 478.53        & 74.39 & \textbf{7.02} & 16.69\\
    2000010.txt     & 439.57        & 117.31        & \textbf{2.18} & 24.14\\
    2cahe10.txt     & 1118.75       & 255.29        & \textbf{29.17}        & 86.71\\
    5wiab10.txt     & 414   & 111.21        & \textbf{4.15} & 25.14\\
    800lg10.txt     & 402.88        & 83.74 & \textbf{3.12} & 16.48\\
    csnva10.txt     & 1721.89       & 107.98        & \textbf{6.86} & 34.72\\
    dbrry10.txt     & 587.39        & 125.46        & \textbf{5.65} & 31.08\\
    dscmn10.txt     & 982.09        & 215.02        & \textbf{19.93}        & 63.73\\
    gltrv10.txt     & 463.34        & 117.35        & \textbf{6.86} & 31.39\\
    milnd10.txt     & 629.47        & 125.02        & \textbf{1.88} & 23.22\\
    mt7bg10.txt     & 1433.58       & 194.1 & \textbf{8.75} & 73.41\\
    stlla10.txt     & 603.45        & 45.14 & 9.67  & \textbf{9.34}\\
    wmcry10.txt     & 592.57        & 143.7 & \textbf{5.25} & 36.47\\
    \midrule
    Mean    & 832.27        & 131.15        & \textbf{8.74} & 36.87\\
    \botrule
  \end{tabular}
\end{table}

We fitted the models defined in (\ref{DConstant})--(\ref{DLogistic})
with the corresponding derived entities described further in \S
\ref{secModels} to the plot of the smoothed number of types $g^*(n)$
defined in (\ref{GNV}). The free parameters of the corresponding hapax
rate models are: $\alpha$ of the offset, $\beta$ of the asymptote, and
$\gamma$ of the scale.  The results of fitting by least squares are
presented in Table \ref{tabPars}.  The quality of the fit can be
witnessed in Table \ref{tabWSSRs}, where WSSR is the weighted sum of
squared residuals and ndf is the number of degrees of freedom, both
computed via the Gnuplot \texttt{fit} command.

\begin{figure}[p]
  \centering
  \vspace{-2em}
  \includegraphics[width=0.8\textwidth]{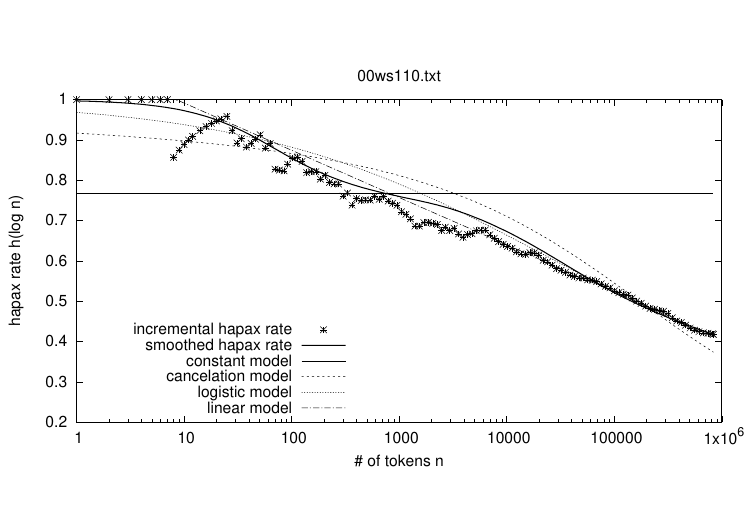}
  \\[-3em]
  \includegraphics[width=0.8\textwidth]{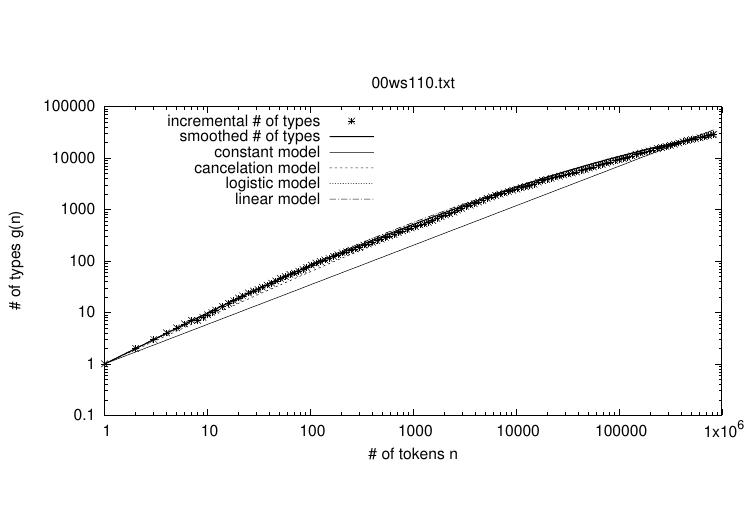}
  \\[-3em]
  \includegraphics[width=0.8\textwidth]{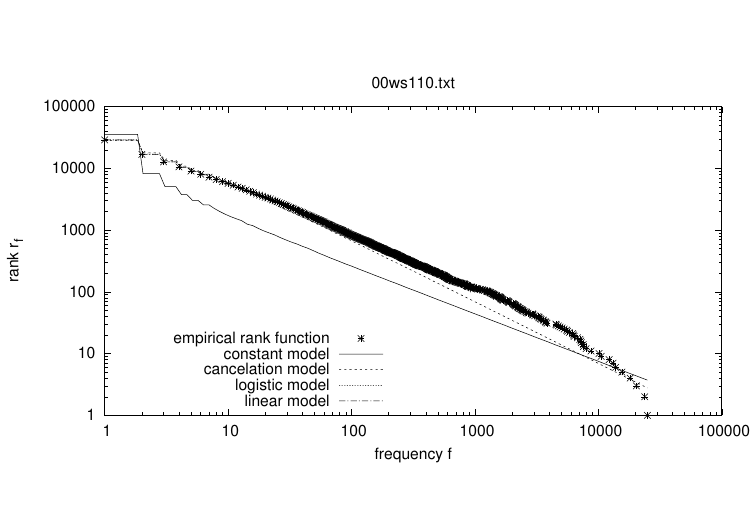}
  \vspace{-2em}
  \caption{W. Shakespeare, \emph{First Folio/35 Plays}.\label{fig00ws110F}}
\end{figure}

\begin{figure}[p]
  \centering
  \vspace{-2em}
  \includegraphics[width=0.8\textwidth]{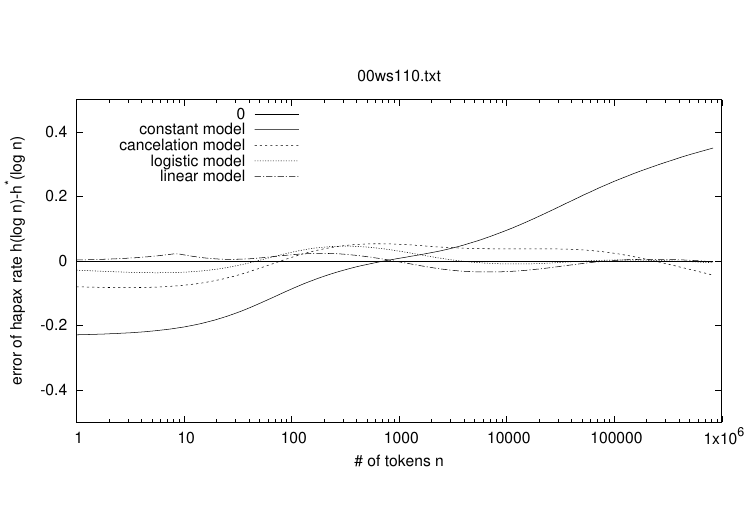}
  \\[-3em]
  \includegraphics[width=0.8\textwidth]{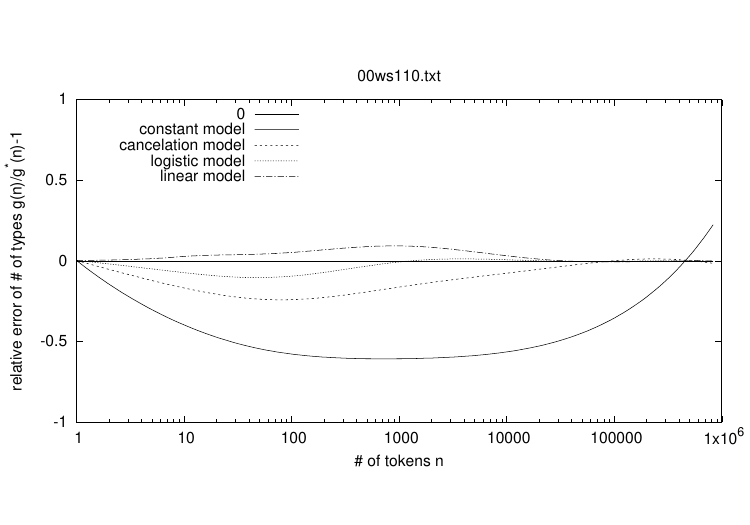}
  \\[-3em]
  \includegraphics[width=0.8\textwidth]{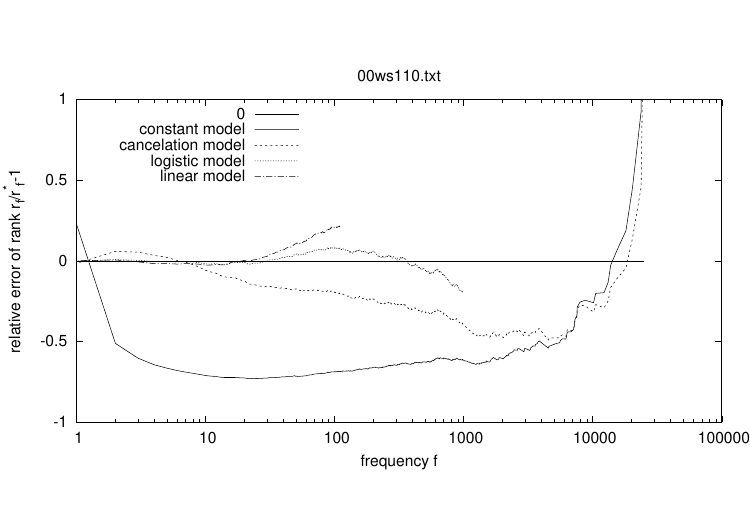}
  \vspace{-2em}
  \caption{W. Shakespeare, \emph{First Folio/35 Plays}.\label{fig00ws110R}}
\end{figure}

Figure \ref{fig00ws110F} contains three plots: the hapax rate, the
vocabulary size (Heaps' law plot), and the rank function (Zipf's law
plot) for \emph{First Folio/35 Plays} by William Shakespeare.  To
visualize the varying quality of particular models in a finer detail,
Figure \ref{fig00ws110R} depicts the fitting residuals for the three
respective plots. The legend to the figures is as follows:
\begin{itemize}
\item The upper plot of Figure \ref{fig00ws110F} shows the hapax rate
  $h(\log n)$ in function of the number of tokens $n$, where the
  incremental hapax rate is $h(u)=h^\#(u)$, the smoothed hapax rate is
  $h(u)=h^*(u)$, the constant model is $h(u)=h_\beta(u)$ given by
  (\ref{DConstant}), the cancelation model is $h(u)=h_\alpha(u)$ given
  by (\ref{DCancelation}), the linear model is
  $h(u)=h_{\alpha\gamma}(u)$ given by (\ref{DLinear}), and the
  logistic model is $h(u)=h_{\alpha\beta\gamma}(u)$ given by
  (\ref{DLogistic}). In the upper plot of Figure \ref{fig00ws110R},
  the error of the hapax rate is defined as $h(u)-h^*(u)$ for the four
  preceding instances of $h(u)$.
\item The middle plot of Figure \ref{fig00ws110F} shows the number of
  types $g(n)$ in function of the number of tokens $n$, where the
  incremental number of types is $g(n)=g^\#(n)$, the smoothed number
  of types is $g(n)=g^*(n)$, the constant model is $g(n)=g_\beta(n)$,
  the cancelation model is $g(n)=g_\alpha(n)$, the linear model is
  $g(n)=g_{\alpha\gamma}(n)$, and the logistic model is
  $g(n)=g_{\alpha\beta\gamma}(n)$.  In the middle plot of Figure
  \ref{fig00ws110R}, the relative error of the number of types is
  defined as $g(n)/g^*(n)-1$ for the four preceding instances of
  $g(n)$.
\item The lower plot of Figure \ref{fig00ws110F} shows the rank $r_f$
  in function of frequency $f$, where the empirical rank function is
  $r_f=r^*_f$, the constant model is $r_f=g_\beta(n||f)$, the
  cancelation model is $r_f=g_\alpha(n||f)$, the linear model is
  $r_f=g_{\alpha\gamma}(n||f)$, and the logistic model is
  $r_f=g_{\alpha\beta\gamma}(n||f)$. In the lower plot of Figure
  \ref{fig00ws110R}, the relative error of the rank is defined as
  $r_f/r^*_f-1$ for the four preceding instances of $r_f$.
\end{itemize}

We notice that in the rank function plots, the logistic model and the
linear model show only the low-frequency portion. This may be due to
either an accumulation of numerical errors in the defining polynomial
recursions or the hypothetical ill-definition of these models, which
we did not manage to exclude on theoretical grounds.  The true reason
is not known yet and further mathematical research is needed to
explain this behavior.

In the supplementary materials we present the analogous graphs for the
remaining texts from Table \ref{tabTexts}. The supplementary scripts
and data are available from Github \citep{Debowski23f}.

\subsection{Discussion}
\label{secDiscussion}

The urn model combined with the analytic function theory provides a
good approximation method for empirical word frequency
distributions. This theoretical framework works well even when it is
combined with simple models of hapax rate decay.  As for the four
considered hapax rate models, for moderately-sized texts from Project
Gutenberg, the constant model is the worst, the cancelation model and
the ill-defined linear model seem better, whereas the logistic model
is usually the best, see Table \ref{tabWSSRs}.  The goodness of fit
between the worst and the best model differs by two orders of
magnitude.

Contrary to the tacit assumption under the Herdan-Heaps law paradigm,
the empirical hapax rate is not constant as a function of the text
length. It exhibits a prominent decaying trend that is approximately
linear in the logarithmic scale over six decades.  By contrast,
Herdan-Heaps' law implies that the hapax rate is constant and almost
twice larger than actually observed for a given whole text. Obviously,
this model is false.  Since the decaying trend of the hapax rate is
approximately linear over many decades then the cancelation model
becomes also suboptimal for sufficiently long texts.

As we have mentioned, the rescaled logistic model is the most accurate
model for novel-sized texts.  In view of the values of parameters
reported in Table \ref{tabPars}, some particularly linguistically
motivated regular case is $\gamma=1/3$ and $\beta=0$. Adjusting
formula (\ref{SLogisticG}), in this case, we obtain the smoothed
number of types of form
\begin{align}
  \label{OneThird}
  g(n)
  =\frac{8n}{(n^{1/3}+1)^3},
\end{align}
which is further rescaled via the offset operation
(\ref{OffsetOne})--(\ref{OffsetTwo}). In other words, with formula
(\ref{OneThird}) and a single parameter $\alpha\approx 10$, we can
reasonably fit all considered texts. Model (\ref{OneThird}) implies
that the number of distinct words is not only asymptotically bounded
by a finite constant but it also takes quite a simple mathematical
form.

\section{Possible extensions}
\label{secMixture}

The improved fitting scheme proposed in this paper should not be
interpreted as a goal by itself but rather as a tool for removing
gross errors to pave a road for better testing of theories of a
linguistic meaning. In particular, there is a hypothesis supported by
\citet{FerrerSole01b}, \citet{MontemurroZanette02},
\citet{PetersenOthers12}, \citet{GerlachAltmann13}, and
\citet{WilliamsOthers15} that natural languages' lexicons consist of a
mixture of a finite core lexicon that is distributed roughly according
to Zipf's law and a periferal lexicon that has a steeper power-law
tail. Moreover, as noticed by \citet{Fan10}, this periferal lexicon,
when extracted from large corpora, contains surprisingly many word
types that do not belong to natural language par excellance, such as
unique random strings.

Combining these two insights, we may consider a ``lexical trash''
hypothesis which states that the periferal lexicon is asymptotically
dominated by monkey-typing and, consequently, it obeys the Mandelbrot
correction of Zipf's law \citep{Mandelbrot54,Miller57}. In the first
approximation, the total lexicon would be a mixture of a finite
roughly Zipf-distributed core lexicon and a monkey-typing
Mandelbrot-corrected periferal lexicon.

To model such a scenario, let us suppose that we have two sets of
candidates for functions $h(u)$ and $g(n)$, namely, $h_i(u)$ and
$g_i(n)$ where $i\in\klam{1,2}$. We observe that these functions can
be combined via the mixture operation
\begin{align}
  h_\lambda(u)&:=\frac{\lambda h_1(u)g_1(e^{u})+(1-\lambda) h_2(u)g_2(e^{u})
                }{\lambda g_1(e^{u})+(1-\lambda) g_2(e^{u})},
  \\
  g_\lambda(n)&:= \lambda g_1(n)+(1-\lambda) g_2(n).
  % \\
  % g_\lambda(n|k)&:= \lambda g_1(n|l)+(1-\lambda) g_2(n|k),
  % \\
  % g_\lambda(n||f)&:= \lambda g_1(n||f)+(1-\lambda) g_2(n||f).
\end{align}
where $\lambda\in(0,1)$ is an empirically chosen parameter. 
% $g(n|k)$ and $g(n||f)$

\begin{figure}[t]
  \centering
  %\vspace{-4em}
  \includegraphics[width=\textwidth]{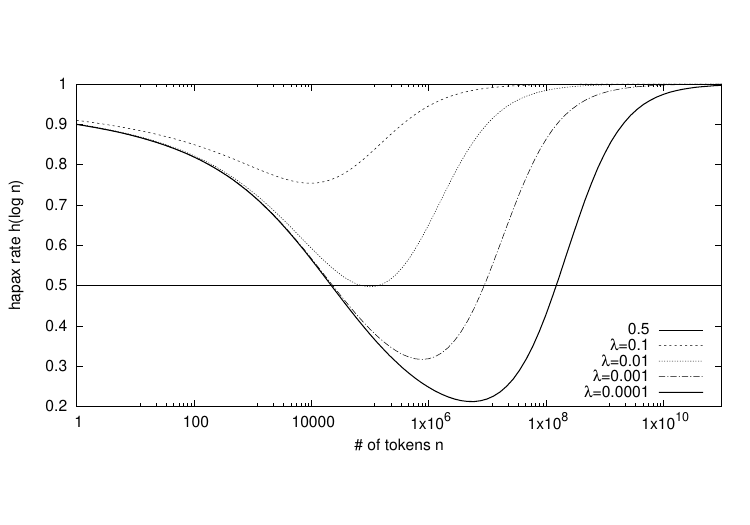}
  %\vspace{-4em}
  \caption{The $U$-shaped hapax rate for a mixture of the cancelation
    model with $\alpha=10$ and the maximal model, i.e., the constant
    model with $\beta=1$. The weight of the cancelation model is
    $(1-\lambda)$ and the weight of the maximal model is
    $\lambda$. \label{figHapaxU}}
\end{figure}

It is interesting to observe that the mixture of a one-parameter
cancelation model (Zipf's law for the core lexicon) and a constant
model (Mandelbrot correction for the periferal lexicon) can easily
produce a $U$-shaped hapax rate plot, see Figure \ref{figHapaxU}.  We
notice that a qualitatively similar $U$-shaped hapax rate plot was
empirically observed for large corpora, see Figures 1 and 2 by
\citet{Fan10}. We also note that, considered as an illustration in
this paper, the sample of 14 texts in English is simply too small to
observe the emergence of the periferal lexicon.

For the space and time constraints of this paper, which has focused on
consolidating the mathematical framework, thorough checking of the
$U$-shaped hapax rate model and the ``lexical trash'' hypothesis is
left for future research. In reality, the intensity of the periferal
lexicon may depend on the degree of cleaning of the investigated
corpus of texts \citet{Fan10} or on factors such as mixing different
sources of texts \citep{WilliamsOthers15}. Thus the meticulous
verification of mixture hapax rate models seems a formidable task,
worth a separate publication.

\section{Conclusion}
\label{secConclusion}

Zipf's and Heaps' laws are celebrated theoretical constructs but their
particular functional forms fail upon a closer scrutiny as too crude
approximation. Borrowing the term from economics, they seem to
constitute stylized facts, i.e., broad tendencies that summarize the
data but ignore the details \citep{Kaldor61}. Thus one might doubt
whether there is a relatively simple but more accurate description of
the word frequency distributions.

Here we have advocated, however, that investigation of the hapax rate
function yields a direct insight where and why particular forms of
Zipf's and Heaps' laws fail. The basic result of this work is to
provide a simple visual explanation of the well-known fact that
Zipf-Mandelbrot and Herdan-Heaps laws are quite inaccurate. This
visual explanation is founded upon seeing in the plot that the
constant model for the hapax rate is clearly wrong. Simultaneously, we
have proposed some new efficiently computable but not completely
obvious corrections to the idealized Zipf-Mandelbrot and Herdan-Heaps
laws. The standard urn model allows to deduce them from a simple
logistic hapax rate model and the predictions are more precise than
for the previously discussed cancelation model.  Our investigation
suggests how to improve on Zipf's and Heaps' laws systematically with
relatively simple parametric models. In particular, we hope that the
hapax rate plot may become a convenient diagnostic tool for future
research in quantitative linguistics.

We suppose that while investigating large natural language corpora, it
may be necessary to consider mixture models where different components
of the lexicon are distinguished based on linguistic criteria and
possibly they obey different hapax rate functions. In particular, we
may empirically research which components of the lexicon are closed or
open---or rather to what degree they are closed or open. Are they
asymptotically bounded (the logistic model with $\beta=0$), do they
grow logarithmically (the cancelation model), or do they follow a
scale-invariant power-law (the constant model)? In fact,
\citet{Baayen01} tried to answer such sort of questions by
investigating productivity of affixes in particular. This work is an
extension of his theoretical ideas into the area of hapax rate
modeling. Our main goal was to provide a relatively self-contained
mathematical toolbox for future research.

\renewcommand{\bibsep}{3pt}
%\bibliography{0-publishers-full,0-journals-full,books,ql,nlp,ai,mine}
\bibliography{zipf_v2}

%\clearpage

\appendix

\section*{Supplementary materials}
  
This is the supplementary report for article ``Corrections of Zipf's
and Heaps' Laws Derived from Hapax Rate Models'', referred to as the
main article.  The supplementary scripts and data are available from
Github \citep{Debowski23f}. In this report, we present the remaining
obtained figures.

\begin{figure}[p]
  \centering
  \vspace{-2em}
  \includegraphics[width=0.8\textwidth]{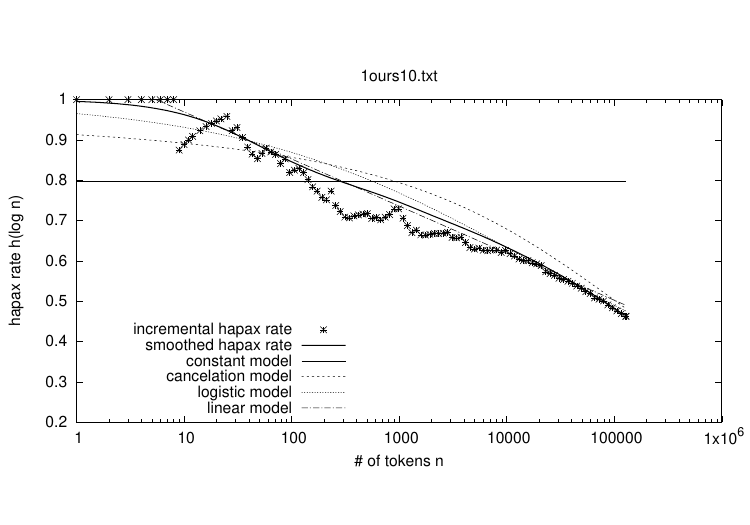}
  \\[-3em]
  \includegraphics[width=0.8\textwidth]{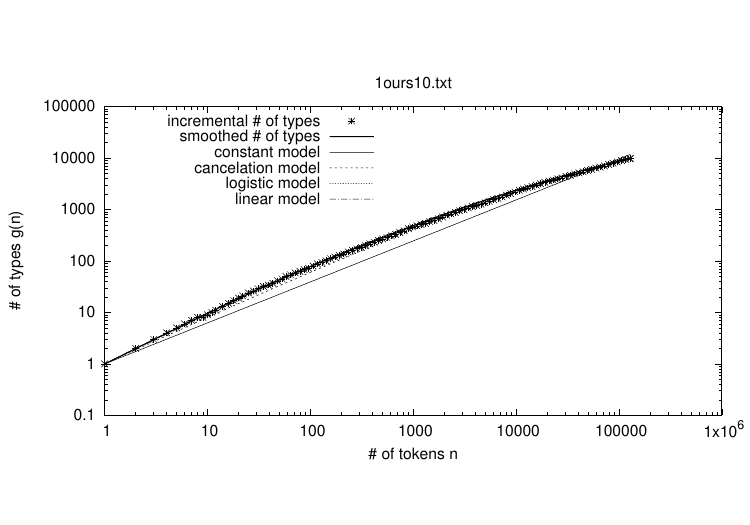}
  \\[-3em]
  \includegraphics[width=0.8\textwidth]{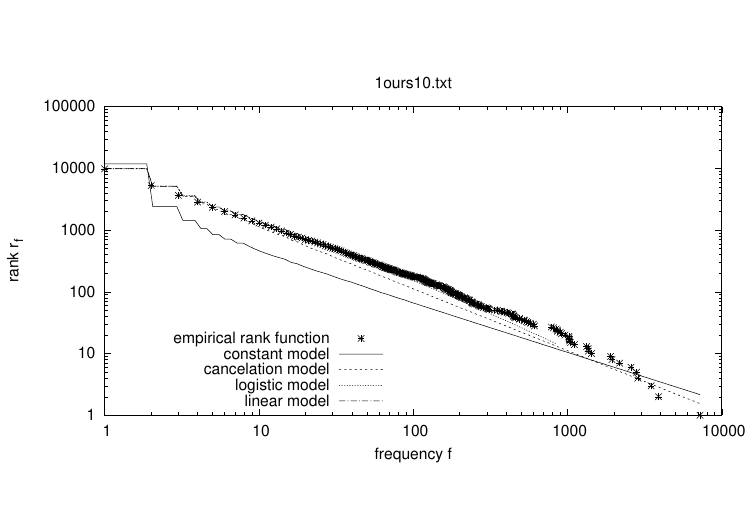}
  \vspace{-2em}
  \caption{W. Cather, \emph{One of Ours}.\label{fig1ours10F}}
\end{figure}

\begin{figure}[p]
  \centering
  \vspace{-2em}
  \includegraphics[width=0.8\textwidth]{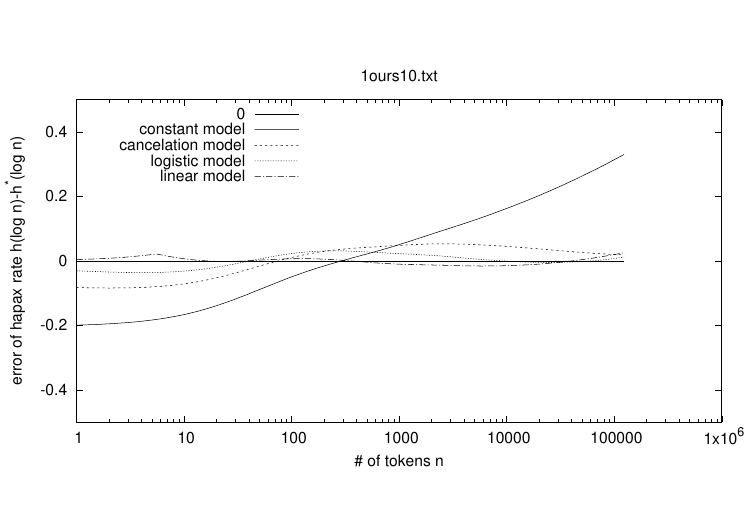}
  \\[-3em]
  \includegraphics[width=0.8\textwidth]{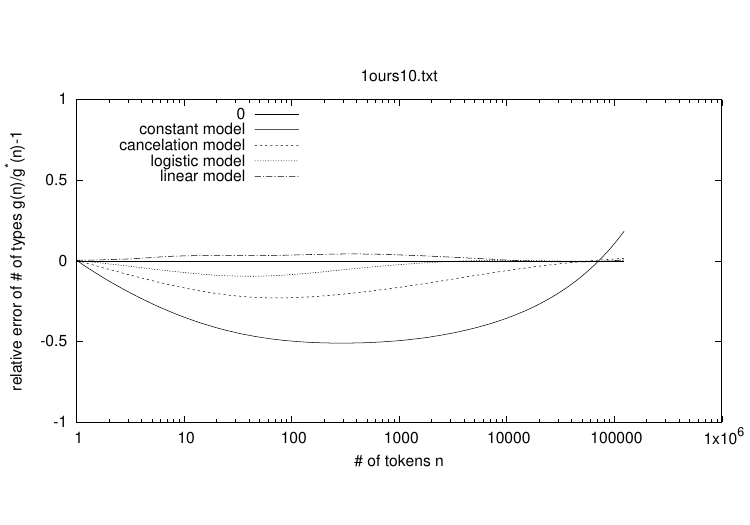}
  \\[-3em]
  \includegraphics[width=0.8\textwidth]{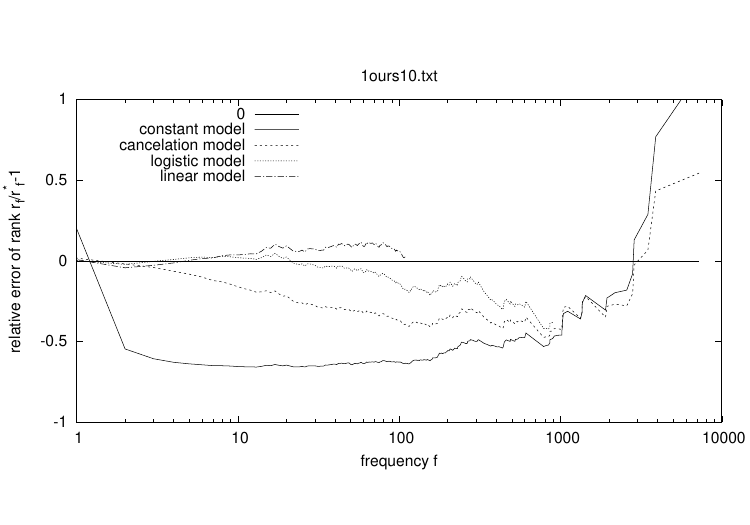}
  \vspace{-2em}
  \caption{W. Cather, \emph{One of Ours}.\label{fig1ours10R}}
\end{figure}

\begin{figure}[p]
  \centering
  \vspace{-2em}
  \includegraphics[width=0.8\textwidth]{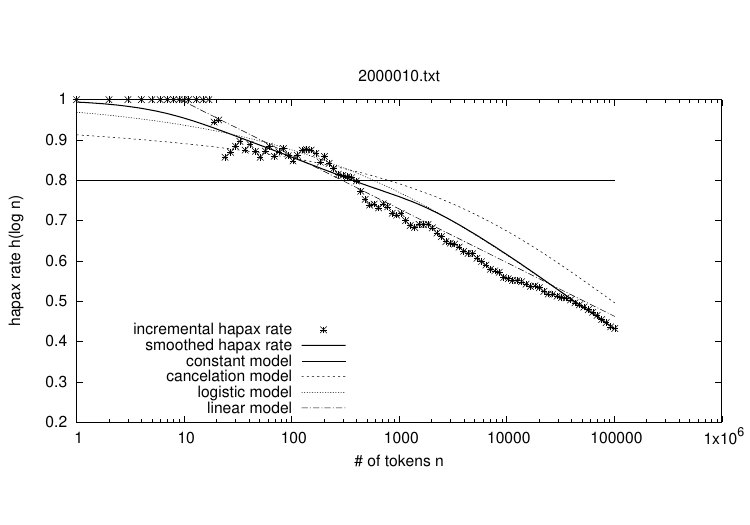}
  \\[-3em]
  \includegraphics[width=0.8\textwidth]{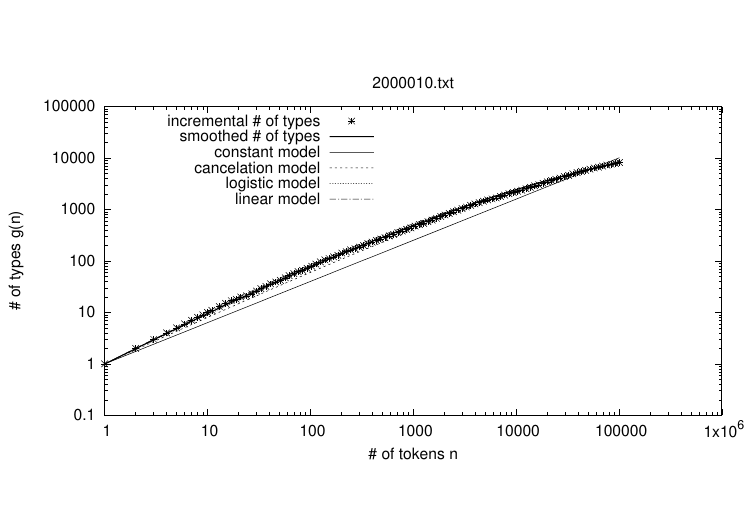}
  \\[-3em]
  \includegraphics[width=0.8\textwidth]{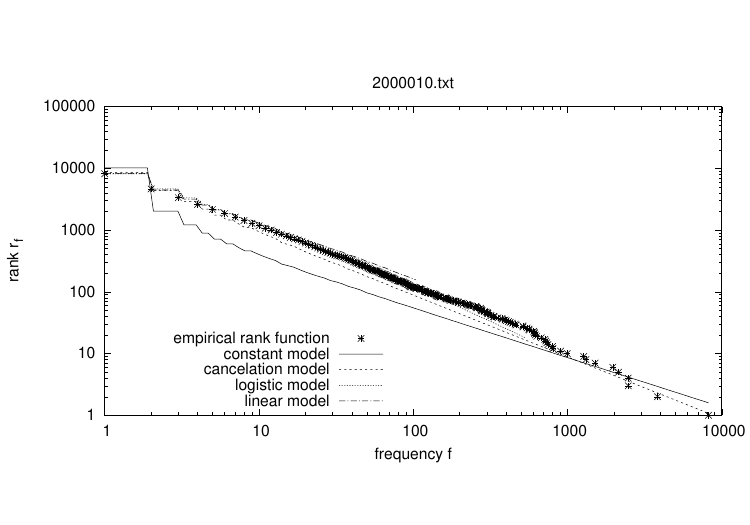}
  \vspace{-2em}
  \caption{J. Verne, \emph{20,000 Leagues under the Sea}.\label{fig2000010F}}
\end{figure}

\begin{figure}[p]
  \centering
  \vspace{-2em}
  \includegraphics[width=0.8\textwidth]{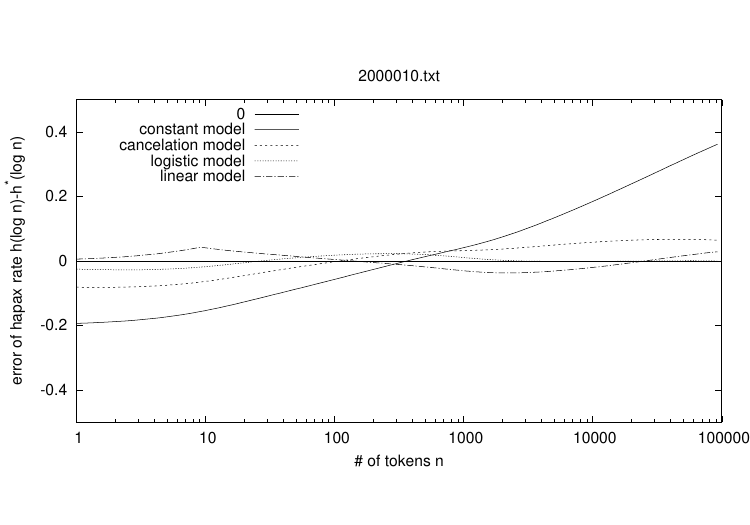}
  \\[-3em]
  \includegraphics[width=0.8\textwidth]{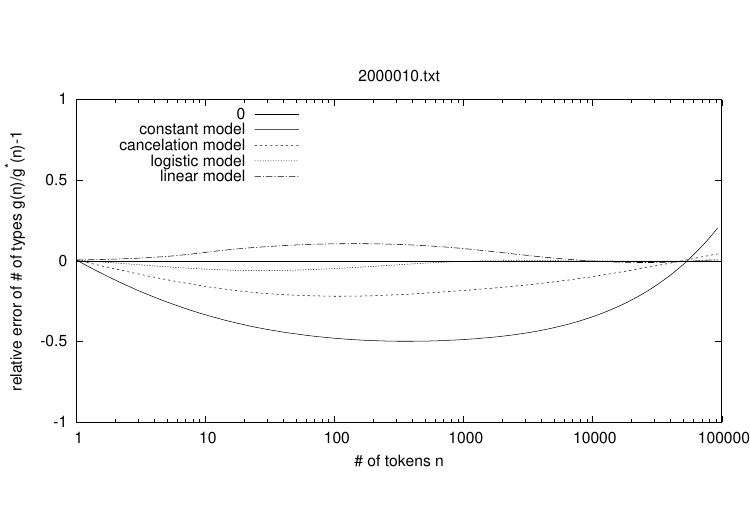}
  \\[-3em]
  \includegraphics[width=0.8\textwidth]{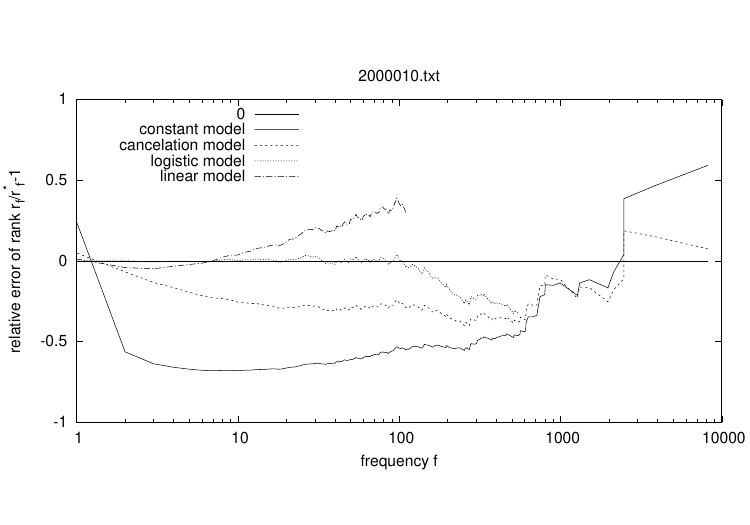}
  \vspace{-2em}
  \caption{J. Verne, \emph{20,000 Leagues under the Sea}.\label{fig2000010R}}
\end{figure}

\begin{figure}[p]
  \centering
  \vspace{-2em}
  \includegraphics[width=0.8\textwidth]{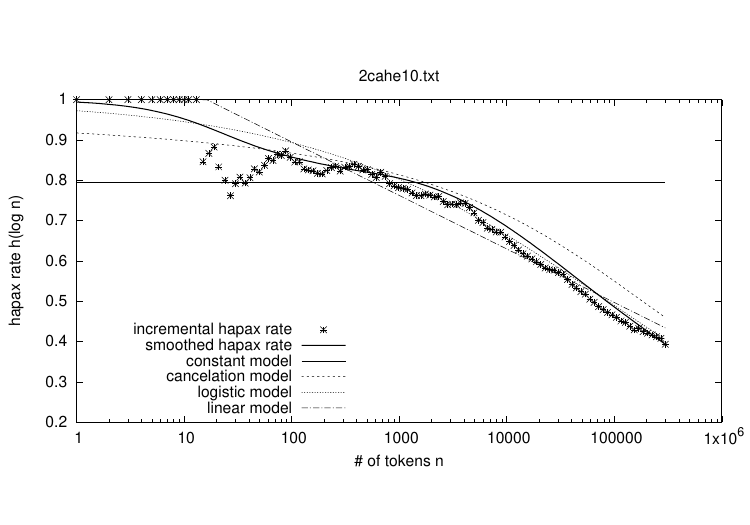}
  \\[-3em]
  \includegraphics[width=0.8\textwidth]{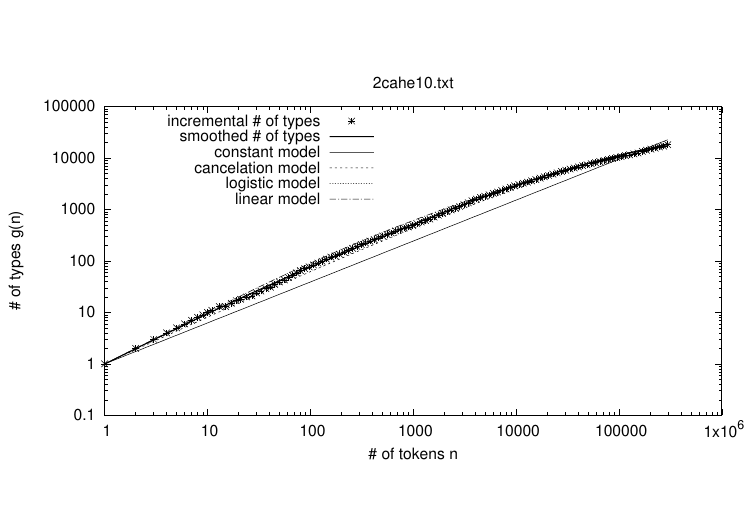}
  \\[-3em]
  \includegraphics[width=0.8\textwidth]{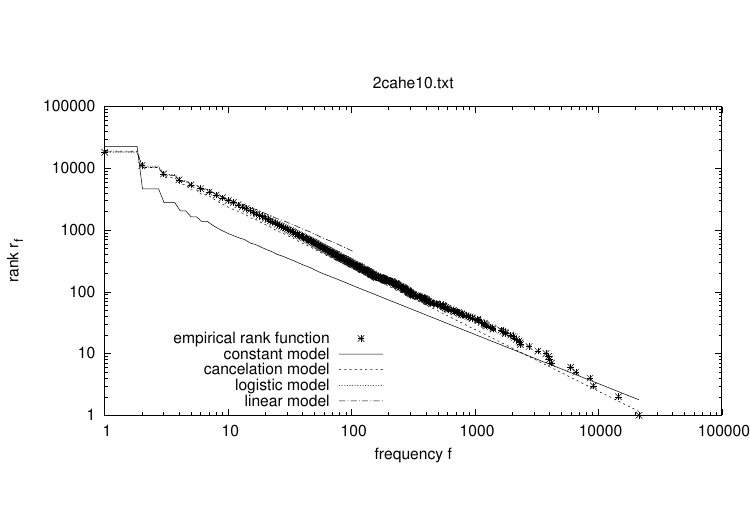}
  \vspace{-2em}
  \caption{T. Macaulay, \emph{Critical \& Historical
      Essays}.\label{fig2cahe10F}}
\end{figure}

\begin{figure}[p]
  \centering
  \vspace{-2em}
  \includegraphics[width=0.8\textwidth]{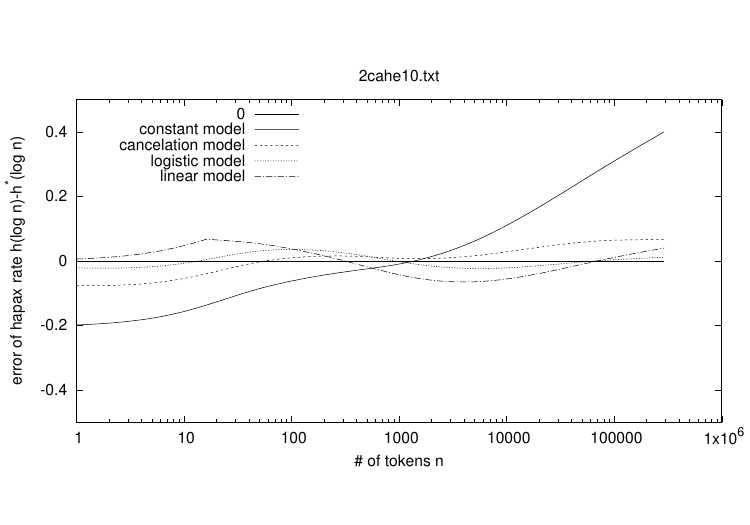}
  \\[-3em]
  \includegraphics[width=0.8\textwidth]{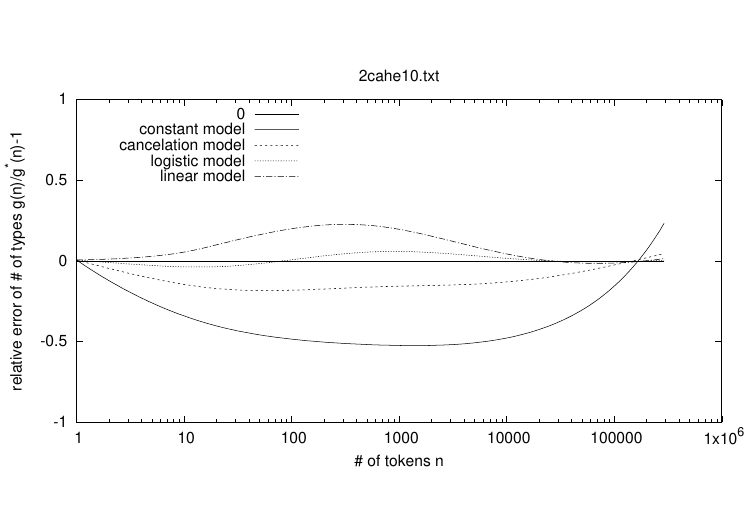}
  \\[-3em]
  \includegraphics[width=0.8\textwidth]{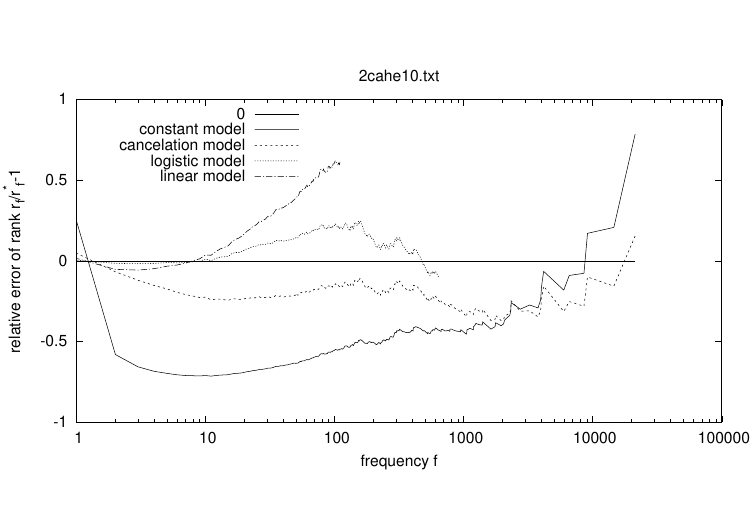}
  \vspace{-2em}
  \caption{T. Macaulay, \emph{Critical \& Historical
      Essays}.\label{fig2cahe10R}}
\end{figure}

\begin{figure}[p]
  \centering
  \vspace{-2em}
  \includegraphics[width=0.8\textwidth]{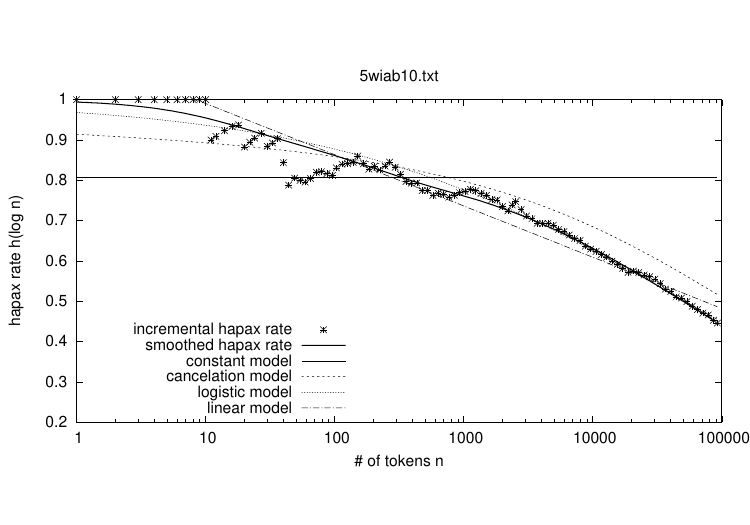}
  \\[-3em]
  \includegraphics[width=0.8\textwidth]{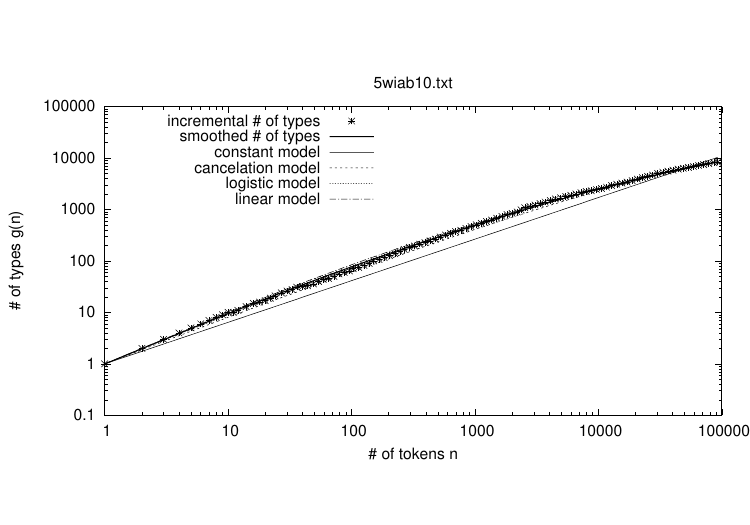}
  \\[-3em]
  \includegraphics[width=0.8\textwidth]{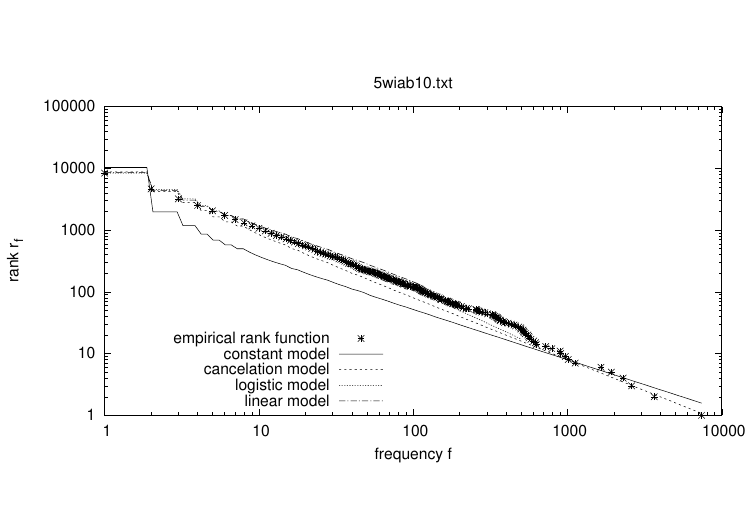}
  \vspace{-2em}
  \caption{J. Verne, \emph{Five Weeks in a Balloon}.\label{fig5wiab10F}}
\end{figure}

\begin{figure}[p]
  \centering
  \vspace{-2em}
  \includegraphics[width=0.8\textwidth]{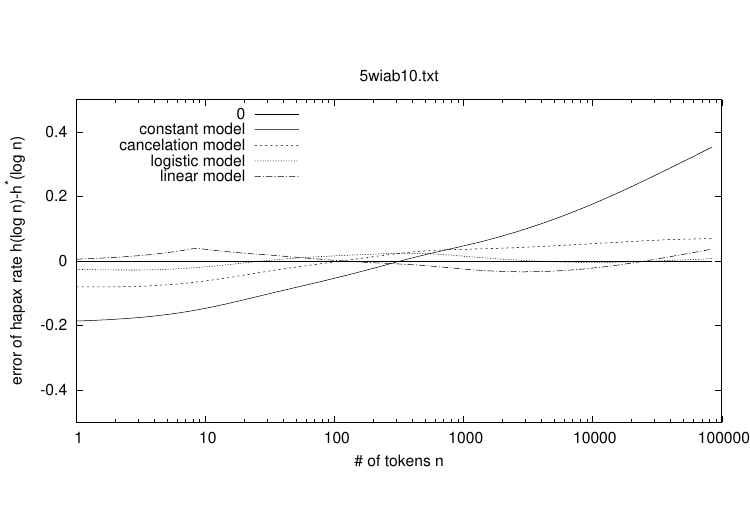}
  \\[-3em]
  \includegraphics[width=0.8\textwidth]{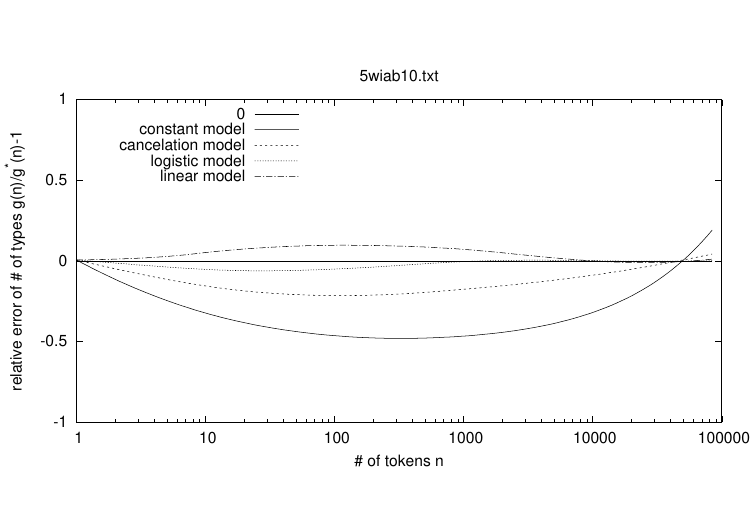}
  \\[-3em]
  \includegraphics[width=0.8\textwidth]{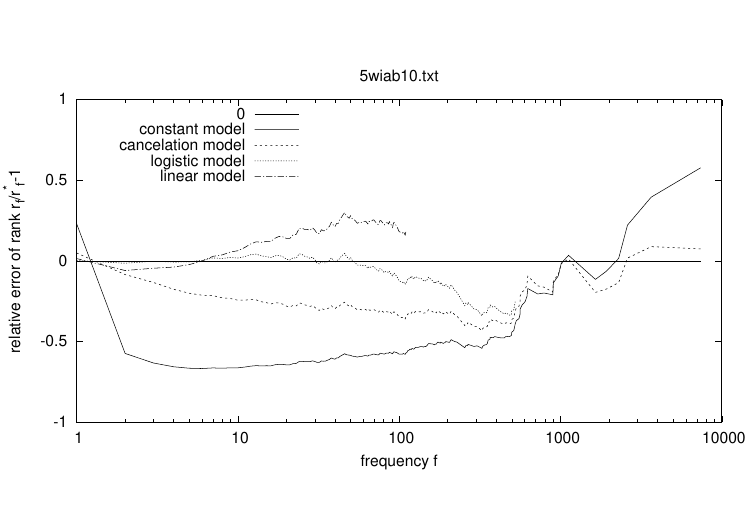}
  \vspace{-2em}
  \caption{J. Verne, \emph{Five Weeks in a Balloon}.\label{fig5wiab10R}}
\end{figure}

\begin{figure}[p]
  \centering
  \vspace{-2em}
  \includegraphics[width=0.8\textwidth]{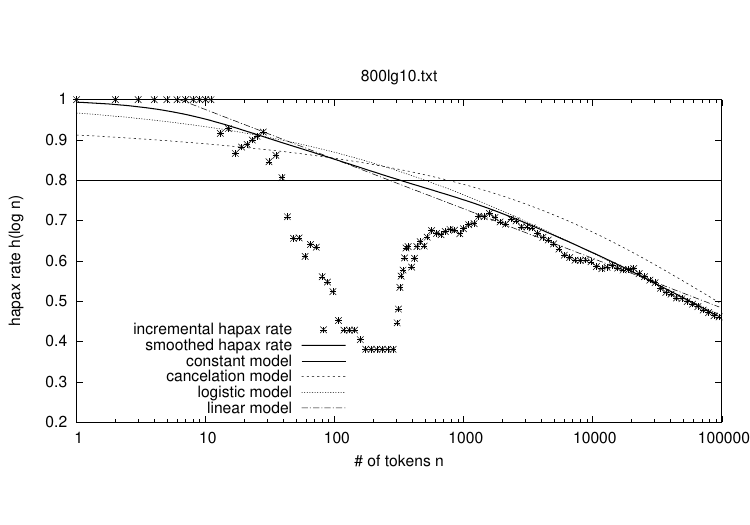}
  \\[-3em]
  \includegraphics[width=0.8\textwidth]{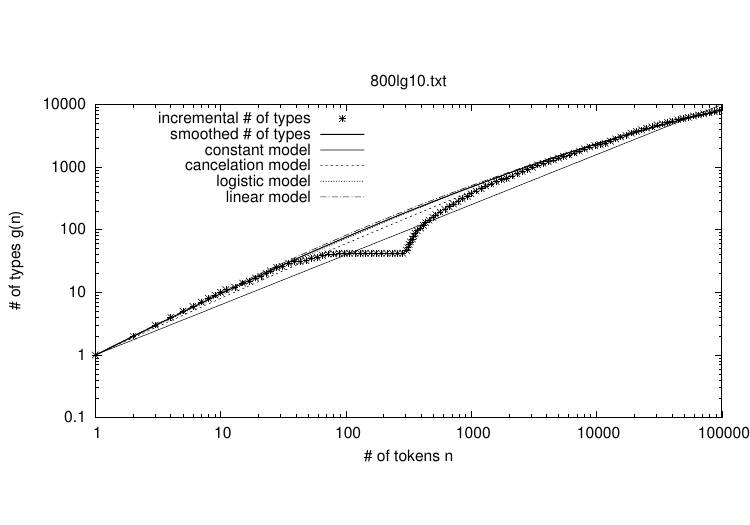}
  \\[-3em]
  \includegraphics[width=0.8\textwidth]{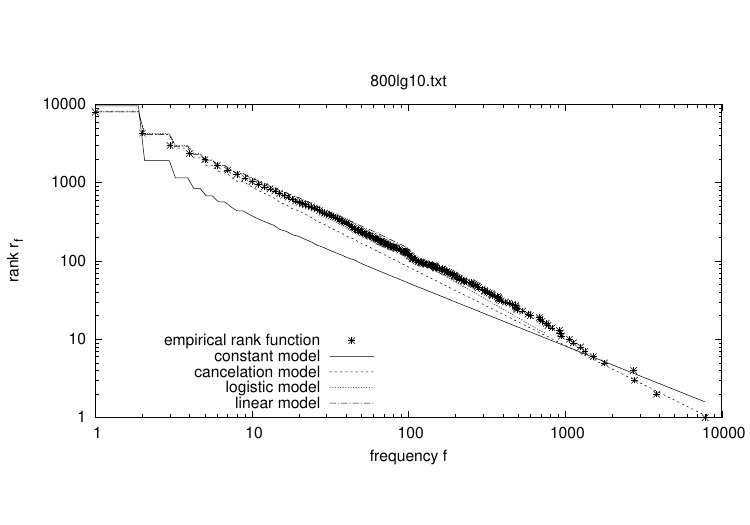}
  \vspace{-2em}
  \caption{J. Verne, \emph{Eight Hundred Leagues on the
      Amazon}.\label{fig800lg10F}}
\end{figure}

\begin{figure}[p]
  \centering
  \vspace{-2em}
  \includegraphics[width=0.8\textwidth]{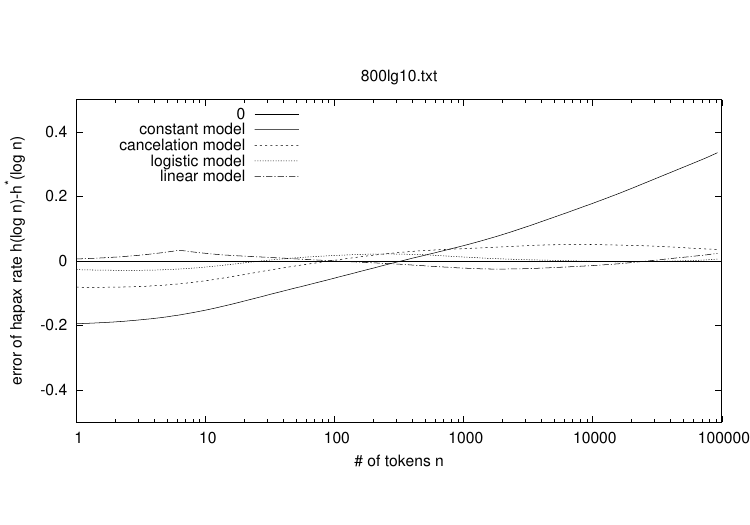}
  \\[-3em]
  \includegraphics[width=0.8\textwidth]{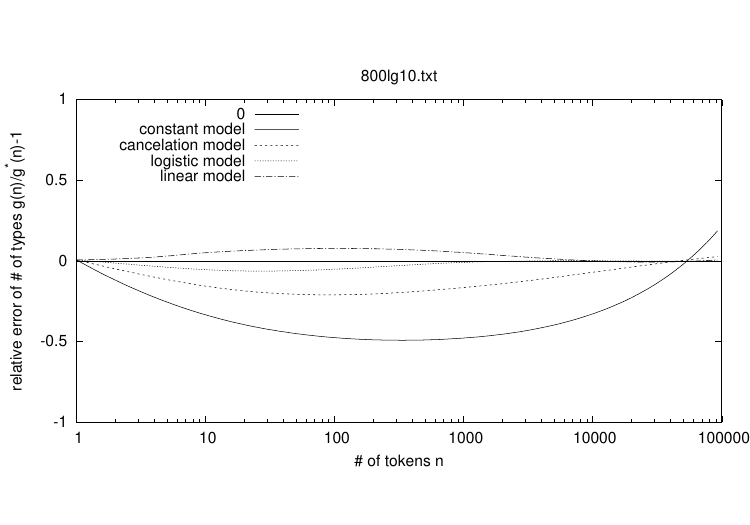}
  \\[-3em]
  \includegraphics[width=0.8\textwidth]{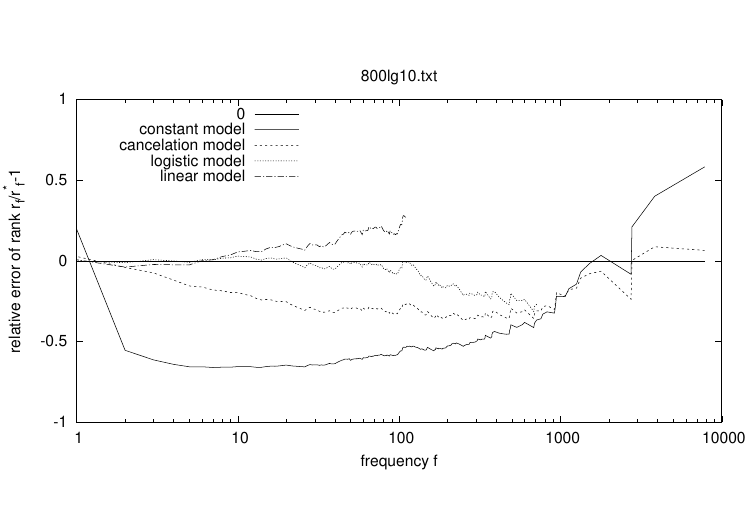}
  \vspace{-2em}
  \caption{J. Verne, \emph{Eight Hundred Leagues on the
      Amazon}.\label{fig800lg10R}}
\end{figure}

\begin{figure}[p]
  \centering
  \vspace{-2em}
  \includegraphics[width=0.8\textwidth]{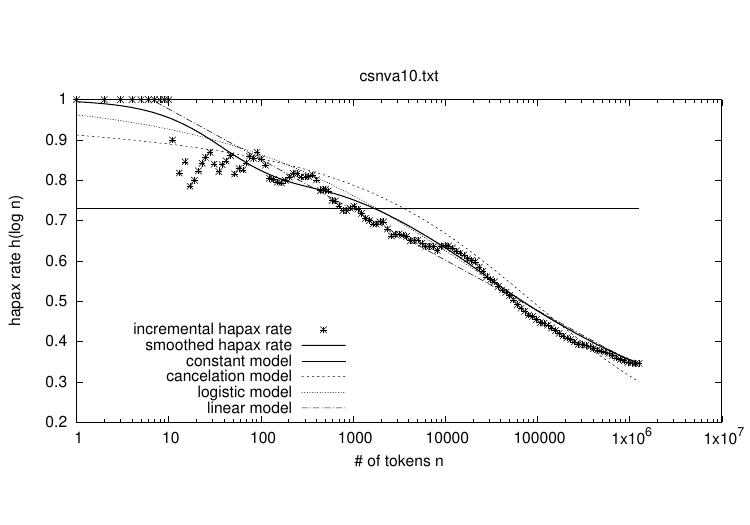}
  \\[-3em]
  \includegraphics[width=0.8\textwidth]{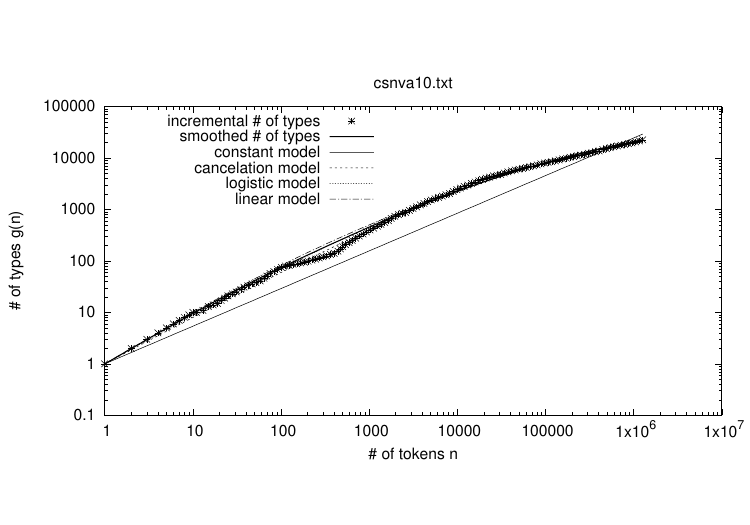}
  \\[-3em]
  \includegraphics[width=0.8\textwidth]{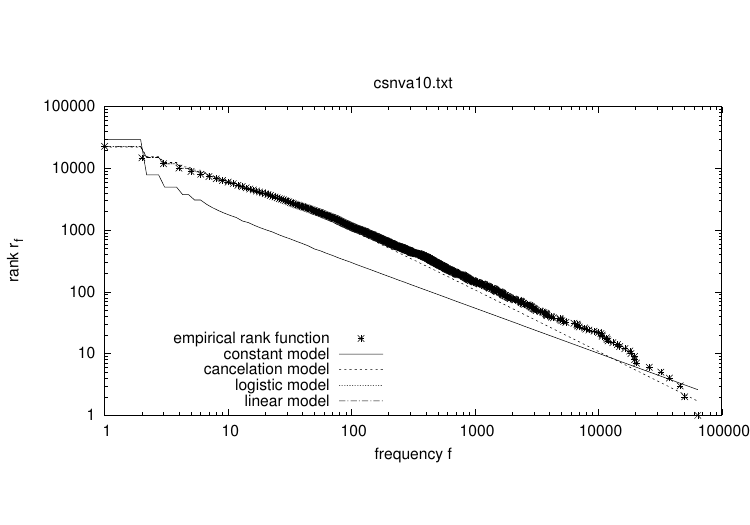}
  \vspace{-2em}
  \caption{J. Casanova, \emph{The Complete Memoirs}.\label{figcsnva10F}}
\end{figure}

\begin{figure}[p]
  \centering
  \vspace{-2em}
  \includegraphics[width=0.8\textwidth]{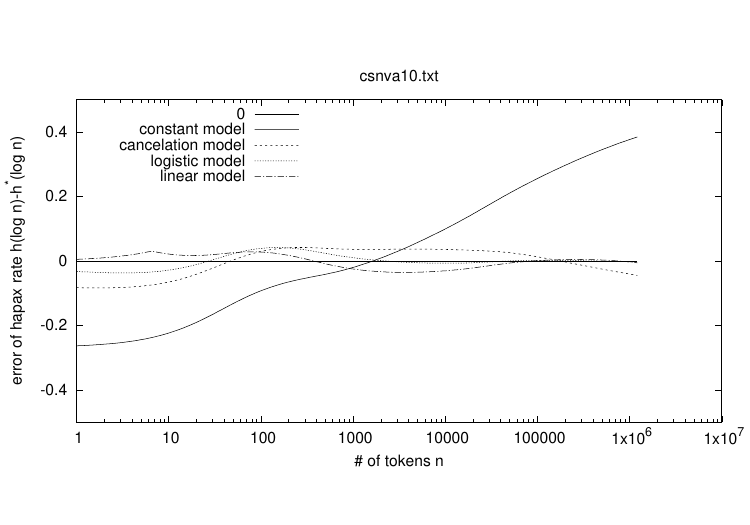}
  \\[-3em]
  \includegraphics[width=0.8\textwidth]{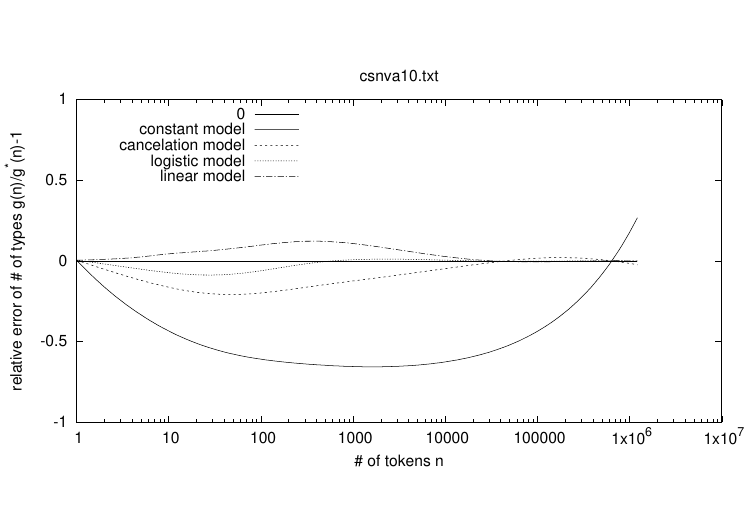}
  \\[-3em]
  \includegraphics[width=0.8\textwidth]{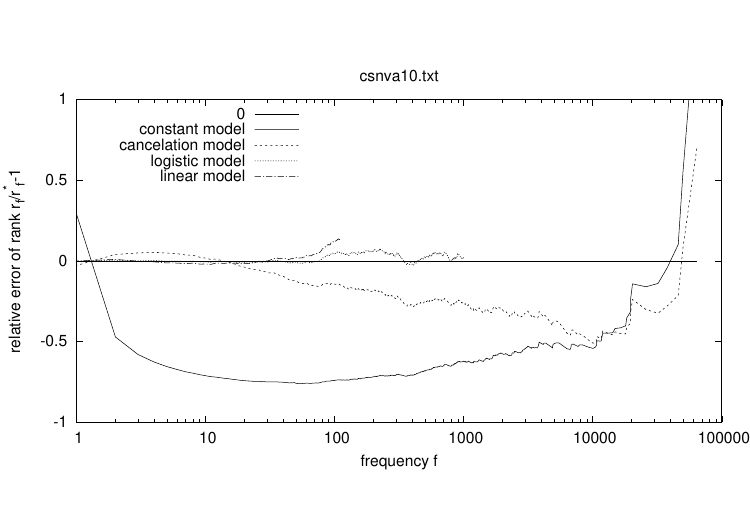}
  \vspace{-2em}
  \caption{J. Casanova, \emph{The Complete Memoirs}.\label{figcsnva10R}}
\end{figure}

\begin{figure}[p]
  \centering
  \vspace{-2em}
  \includegraphics[width=0.8\textwidth]{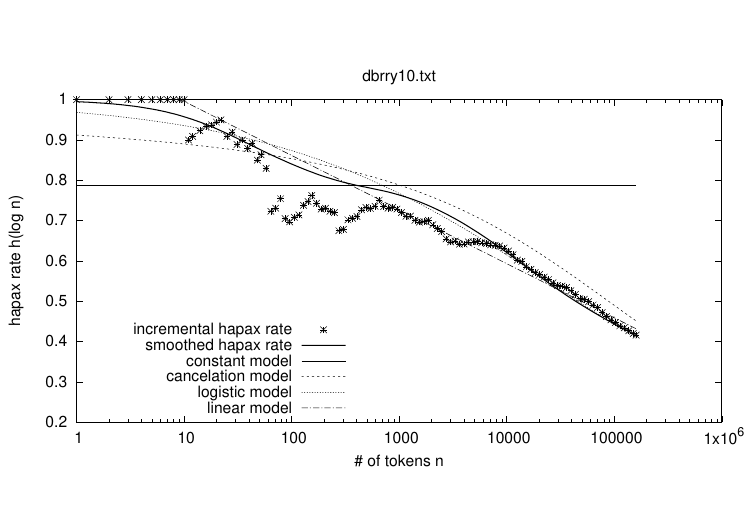}
  \\[-3em]
  \includegraphics[width=0.8\textwidth]{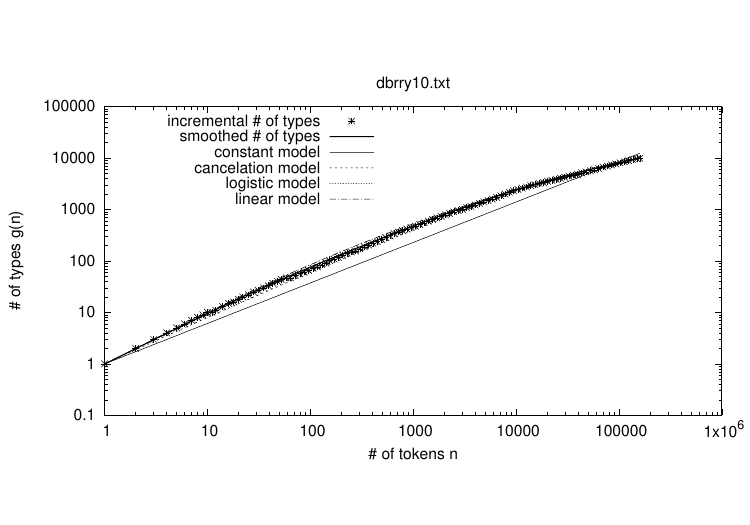}
  \\[-3em]
  \includegraphics[width=0.8\textwidth]{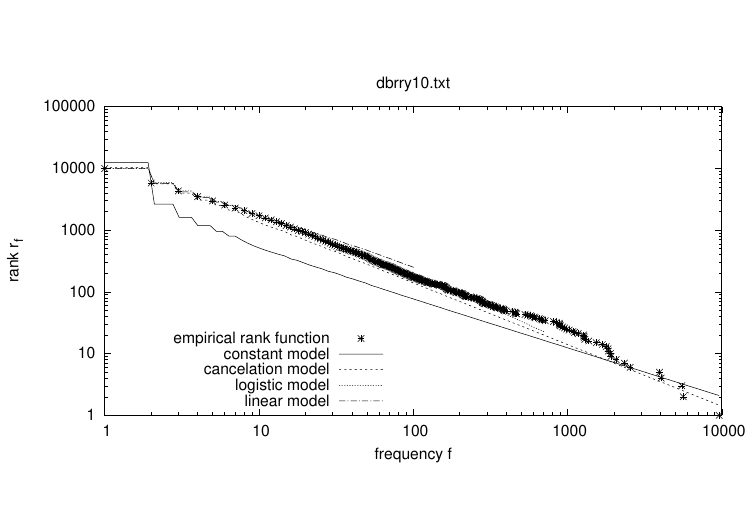}
  \vspace{-2em}
  \caption{Comtesse du Barry, \emph{Memoirs}.\label{figdbrry10F}}
\end{figure}

\begin{figure}[p]
  \centering
  \vspace{-2em}
  \includegraphics[width=0.8\textwidth]{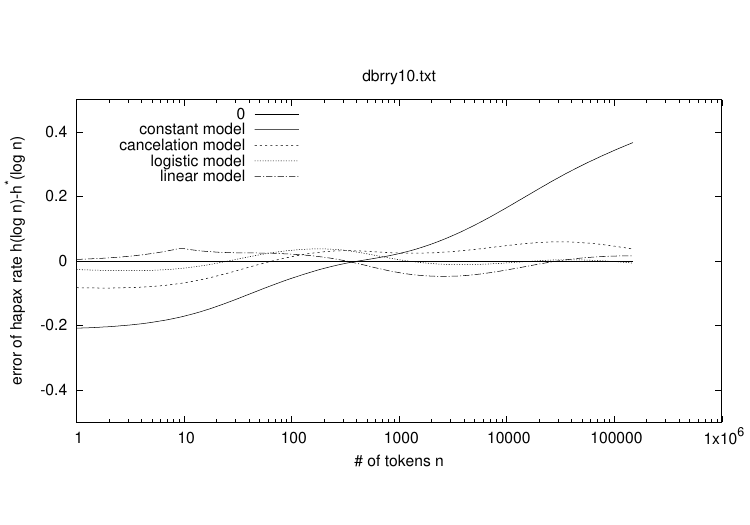}
  \\[-3em]
  \includegraphics[width=0.8\textwidth]{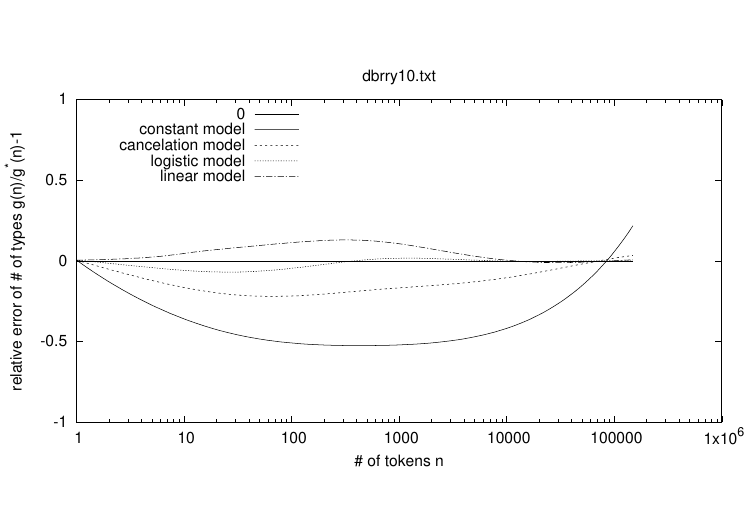}
  \\[-3em]
  \includegraphics[width=0.8\textwidth]{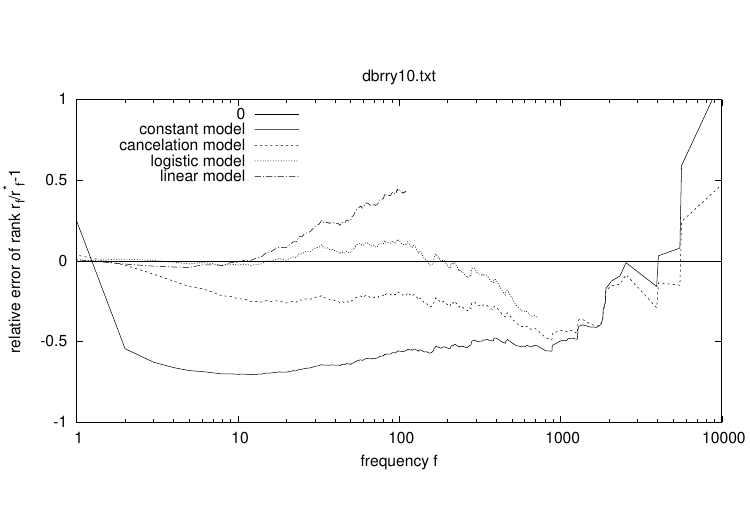}
  \vspace{-2em}
  \caption{Comtesse du Barry, \emph{Memoirs}.\label{figdbrry10R}}
\end{figure}

\begin{figure}[p]
  \centering
  \vspace{-2em}
  \includegraphics[width=0.8\textwidth]{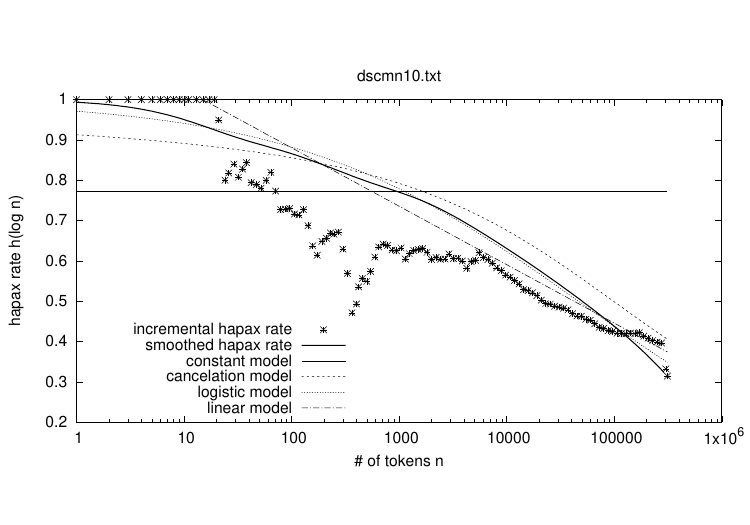}
  \\[-3em]
  \includegraphics[width=0.8\textwidth]{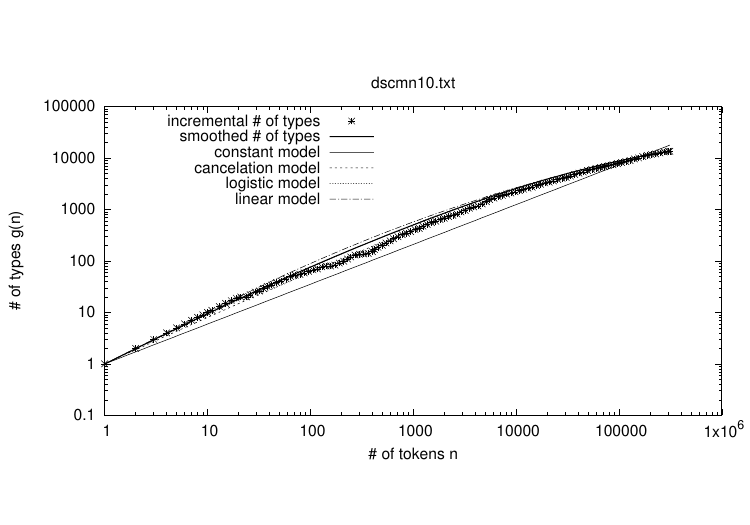}
  \\[-3em]
  \includegraphics[width=0.8\textwidth]{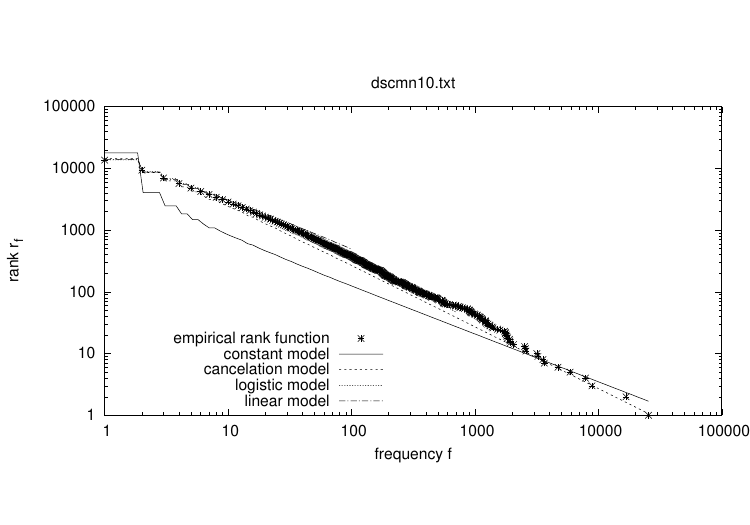}
  \vspace{-2em}
  \caption{C. Darwin, \emph{The Descent of Man}.\label{figdscmn10F}}
\end{figure}

\begin{figure}[p]
  \centering
  \vspace{-2em}
  \includegraphics[width=0.8\textwidth]{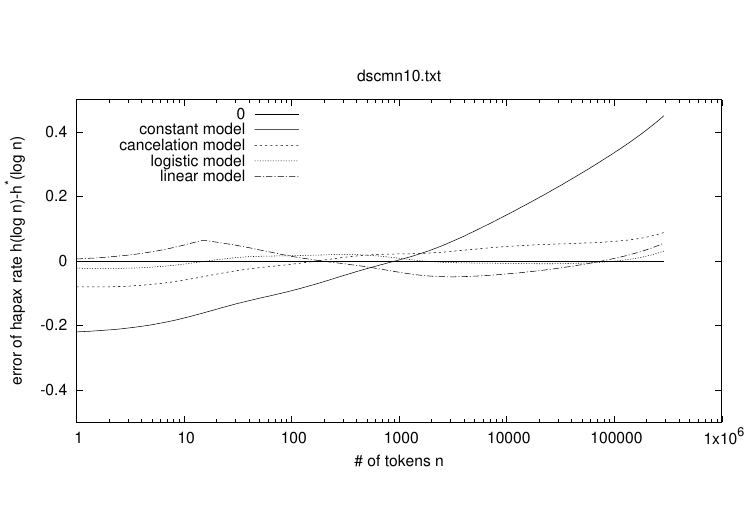}
  \\[-3em]
  \includegraphics[width=0.8\textwidth]{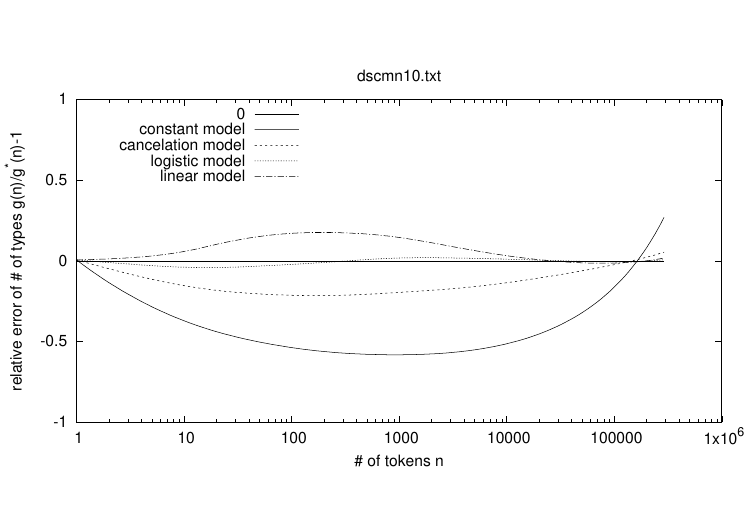}
  \\[-3em]
  \includegraphics[width=0.8\textwidth]{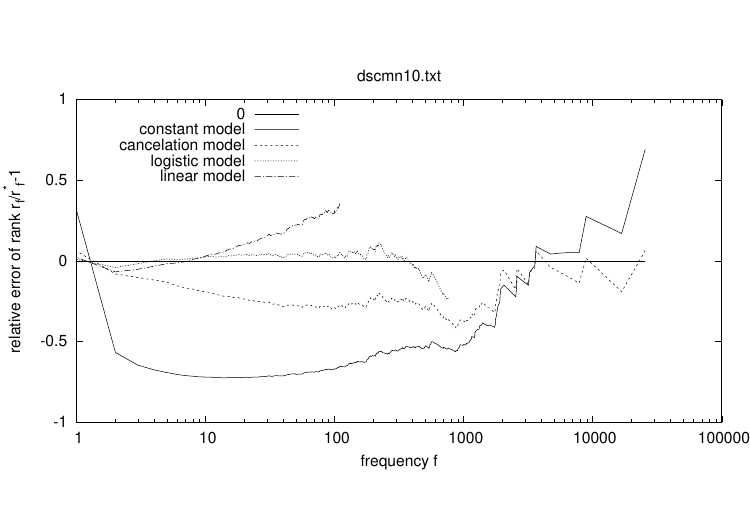}
  \vspace{-2em}
  \caption{C. Darwin, \emph{The Descent of Man}.\label{figdscmn10R}}
\end{figure}

\begin{figure}[p]
  \centering
  \vspace{-2em}
  \includegraphics[width=0.8\textwidth]{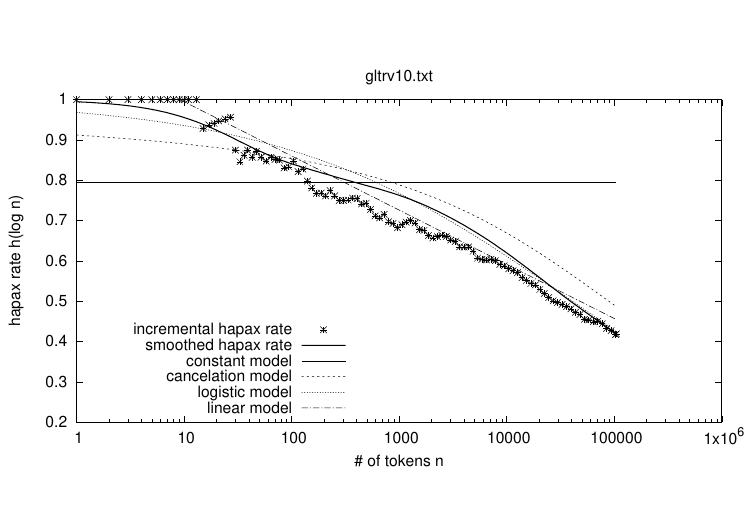}
  \\[-3em]
  \includegraphics[width=0.8\textwidth]{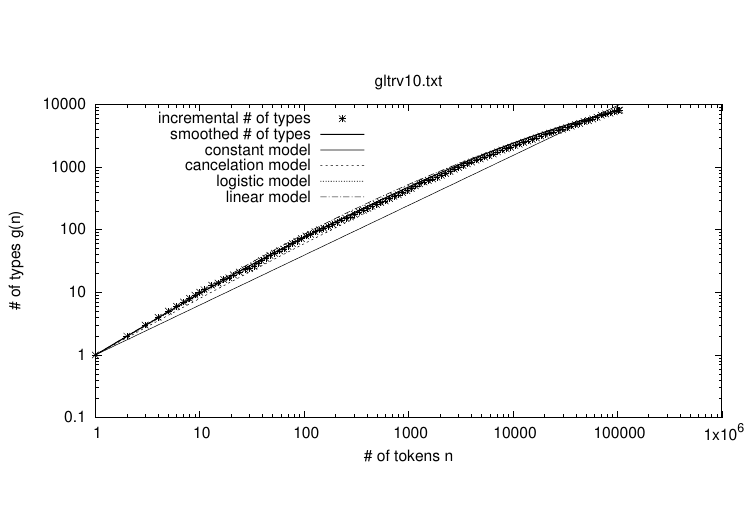}
  \\[-3em]
  \includegraphics[width=0.8\textwidth]{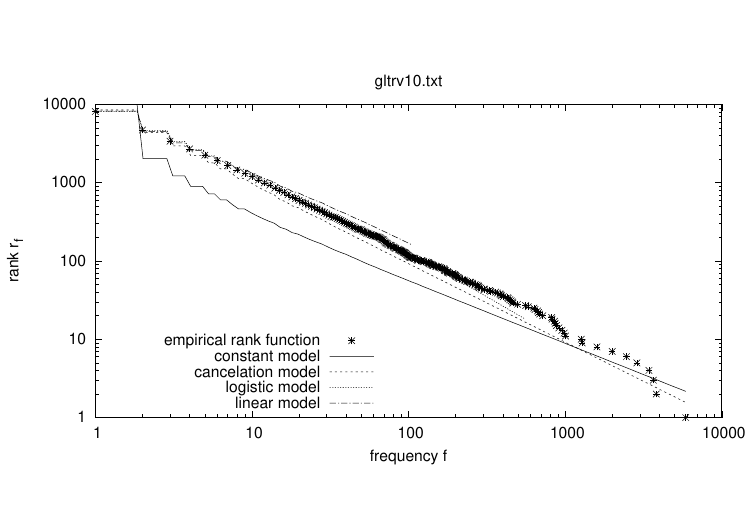}
  \vspace{-2em}
  \caption{J. Swift, \emph{Gulliver's Travels}.\label{figgltrv10F}}
\end{figure}

\begin{figure}[p]
  \centering
  \vspace{-2em}
  \includegraphics[width=0.8\textwidth]{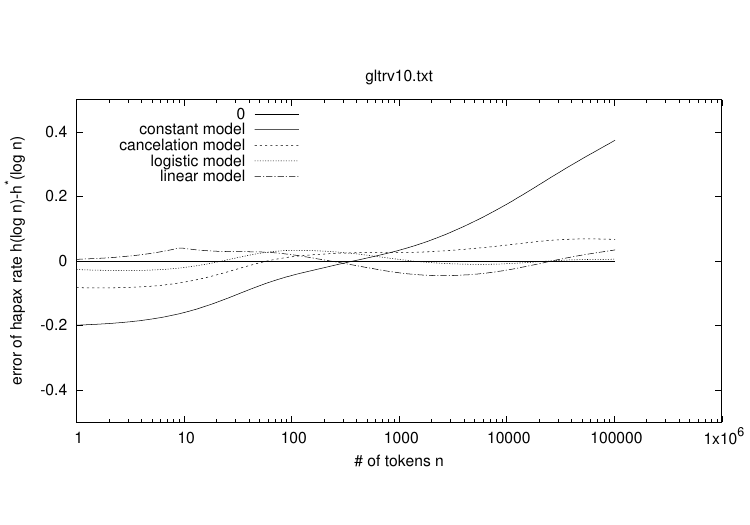}
  \\[-3em]
  \includegraphics[width=0.8\textwidth]{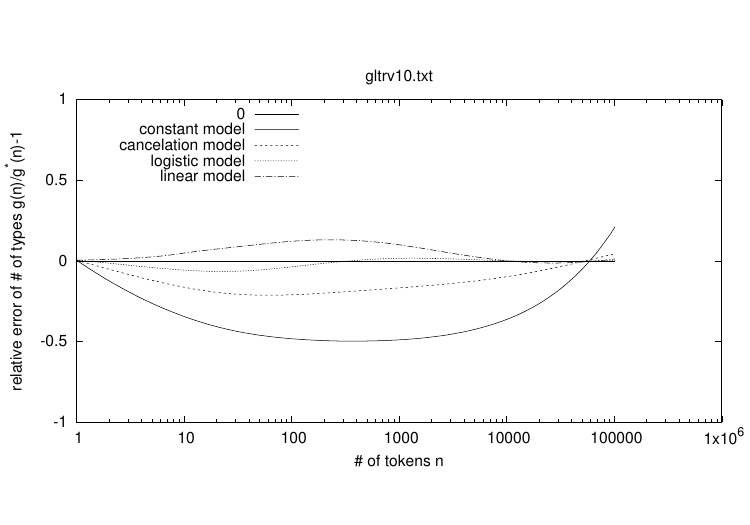}
  \\[-3em]
  \includegraphics[width=0.8\textwidth]{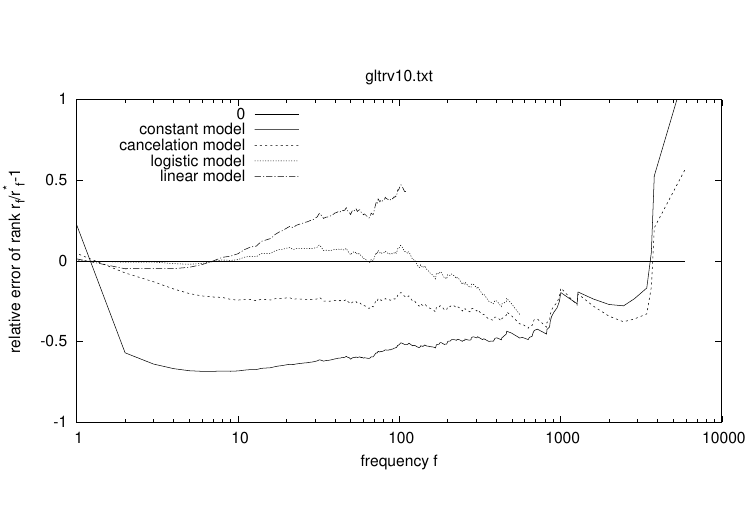}
  \vspace{-2em}
  \caption{J. Swift, \emph{Gulliver's Travels}.\label{figgltrv10R}}
\end{figure}

\begin{figure}[p]
  \centering
  \vspace{-2em}
  \includegraphics[width=0.8\textwidth]{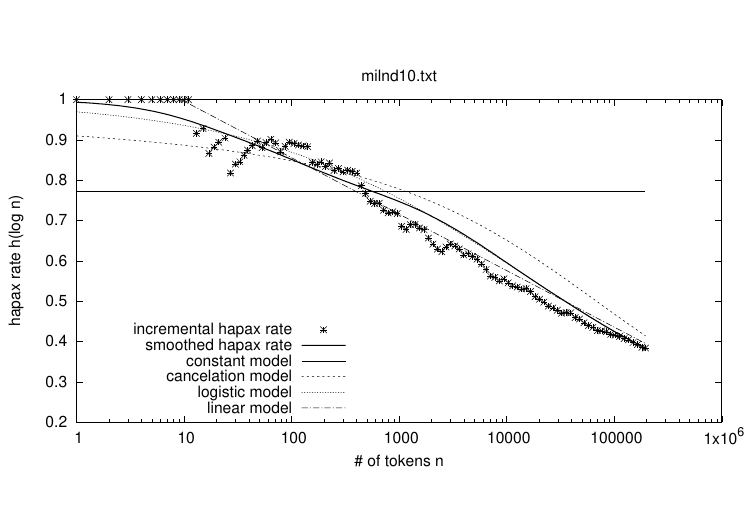}
  \\[-3em]
  \includegraphics[width=0.8\textwidth]{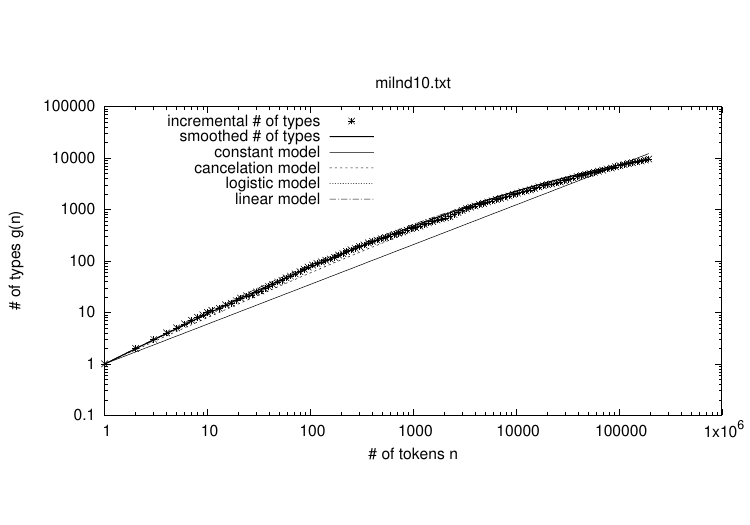}
  \\[-3em]
  \includegraphics[width=0.8\textwidth]{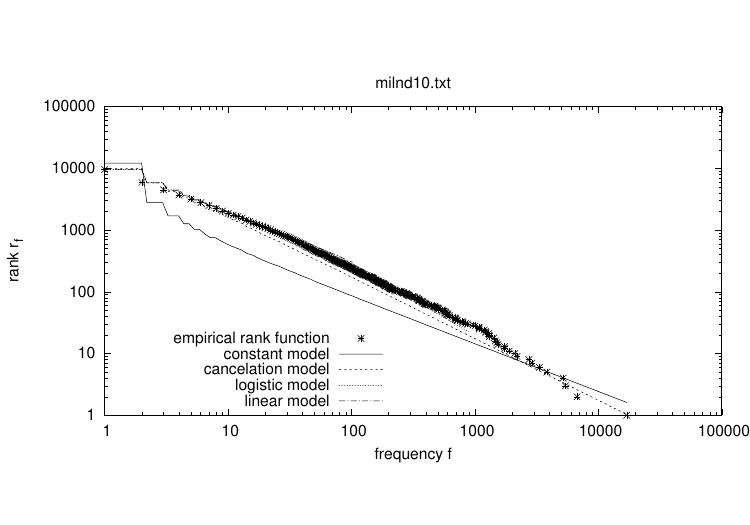}
  \vspace{-2em}
  \caption{J. Verne, \emph{The Mysterious Island}.\label{figmilnd10F}}
\end{figure}

\begin{figure}[p]
  \centering
  \vspace{-2em}
  \includegraphics[width=0.8\textwidth]{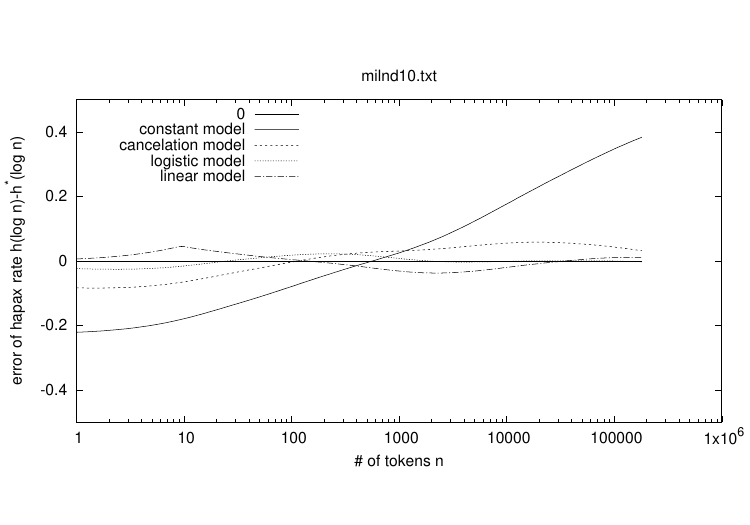}
  \\[-3em]
  \includegraphics[width=0.8\textwidth]{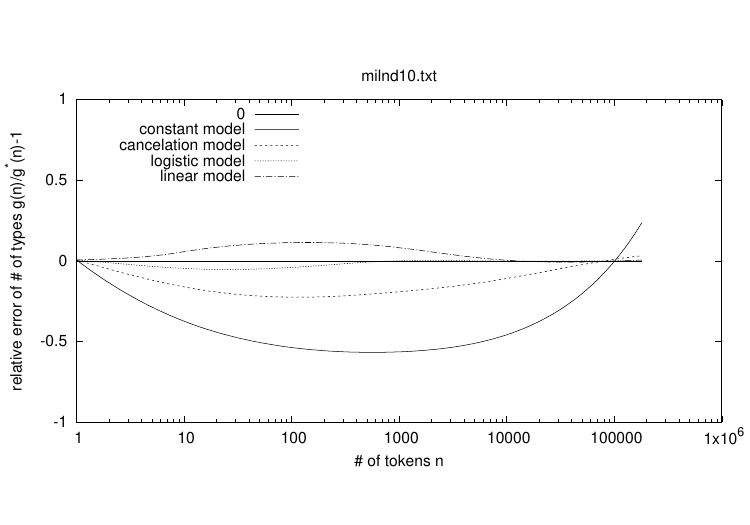}
  \\[-3em]
  \includegraphics[width=0.8\textwidth]{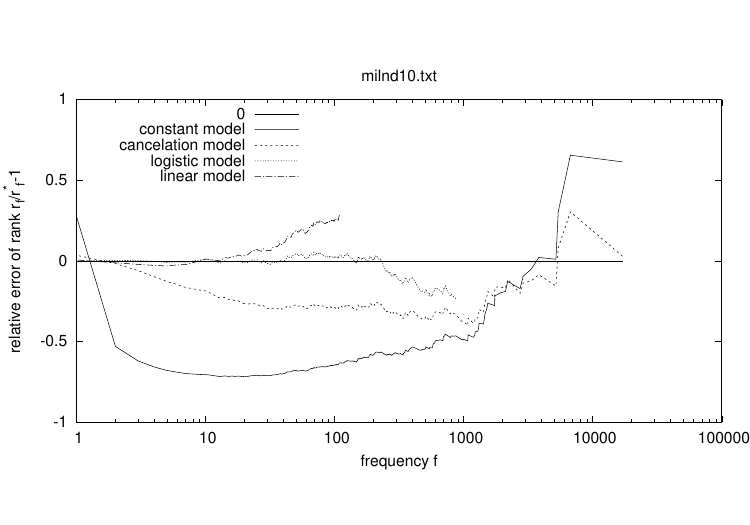}
  \vspace{-2em}
  \caption{J. Verne, \emph{The Mysterious Island}.\label{figmilnd10R}}
\end{figure}

\begin{figure}[p]
  \centering
  \vspace{-2em}
  \includegraphics[width=0.8\textwidth]{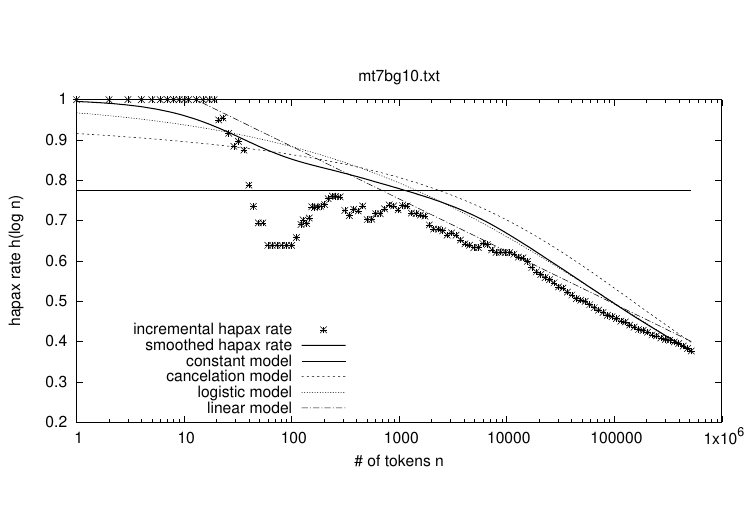}
  \\[-3em]
  \includegraphics[width=0.8\textwidth]{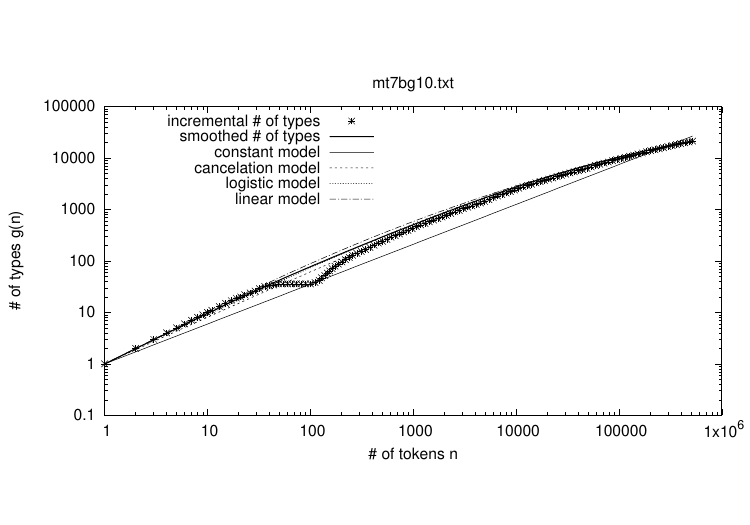}
  \\[-3em]
  \includegraphics[width=0.8\textwidth]{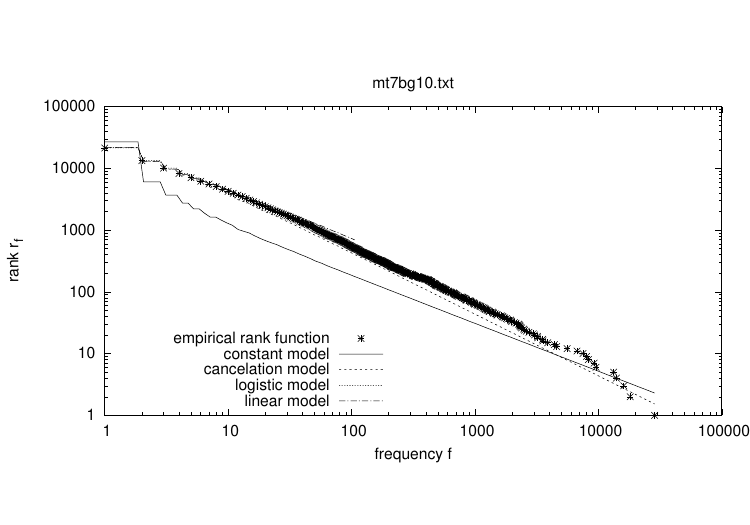}
  \vspace{-2em}
  \caption{A. Paine, \emph{Mark Twain, A Biography}.\label{figmt7bg10F}}
\end{figure}

\begin{figure}[p]
  \centering
  \vspace{-2em}
  \includegraphics[width=0.8\textwidth]{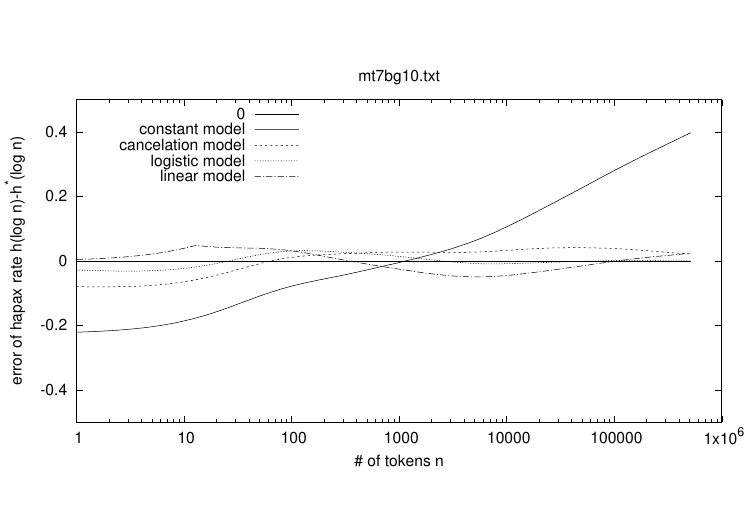}
  \\[-3em]
  \includegraphics[width=0.8\textwidth]{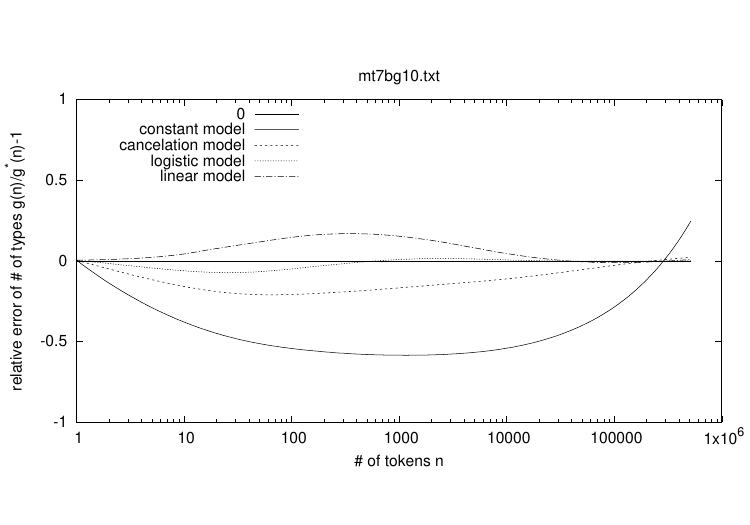}
  \\[-3em]
  \includegraphics[width=0.8\textwidth]{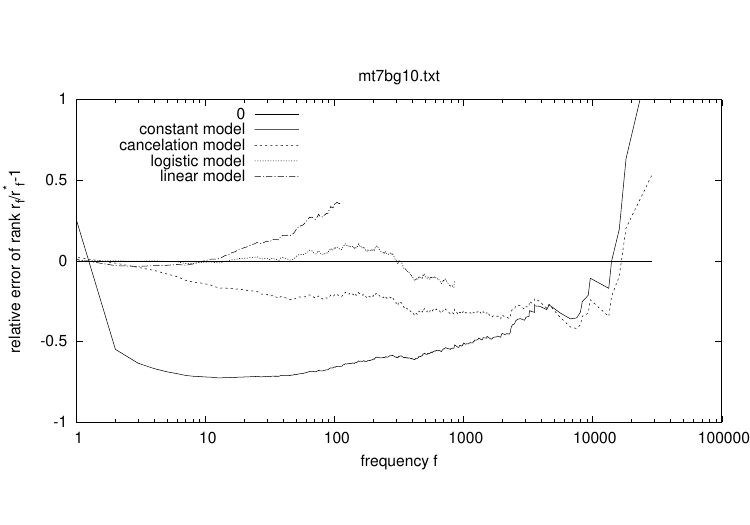}
  \vspace{-2em}
  \caption{A. Paine, \emph{Mark Twain, A Biography}.\label{figmt7bg10R}}
\end{figure}

\begin{figure}[p]
  \centering
  \vspace{-2em}
  \includegraphics[width=0.8\textwidth]{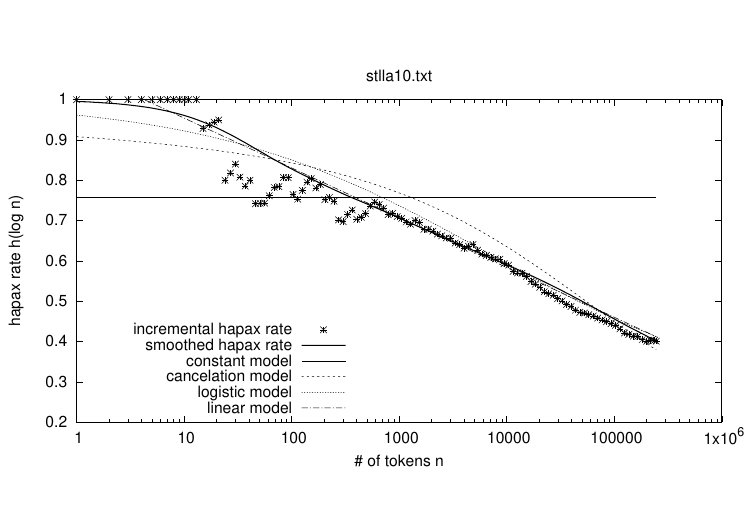}
  \\[-3em]
  \includegraphics[width=0.8\textwidth]{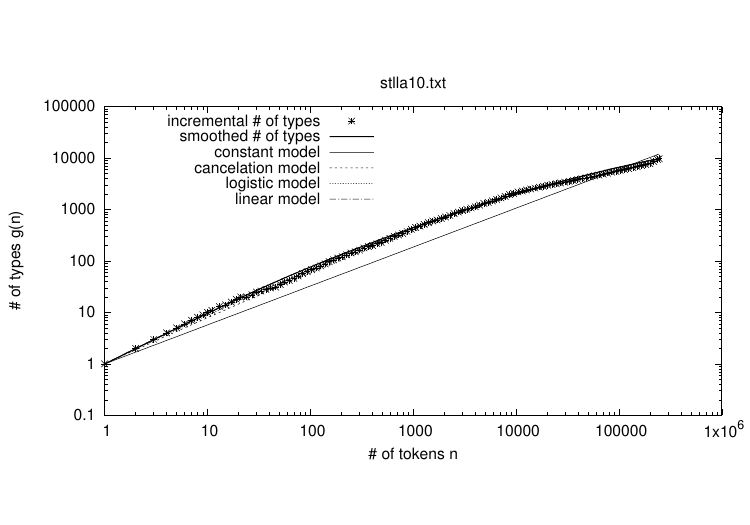}
  \\[-3em]
  \includegraphics[width=0.8\textwidth]{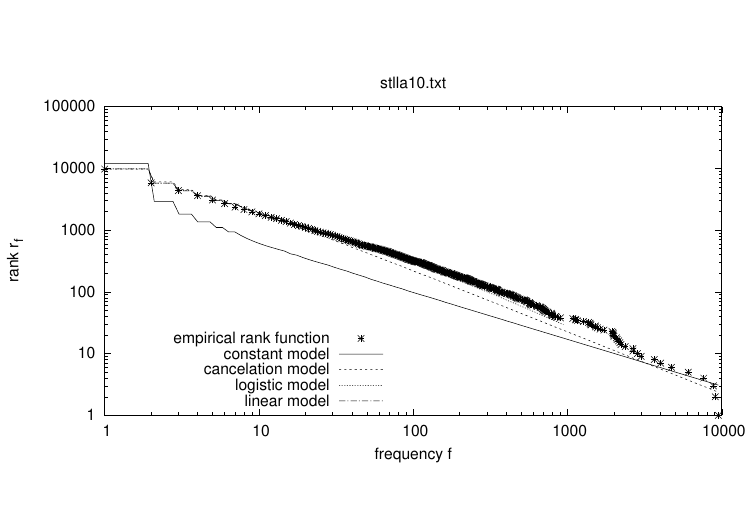}
  \vspace{-2em}
  \caption{J. Swift, \emph{The Journal to Stella}.\label{figstlla10F}}
\end{figure}

\begin{figure}[p]
  \centering
  \vspace{-2em}
  \includegraphics[width=0.8\textwidth]{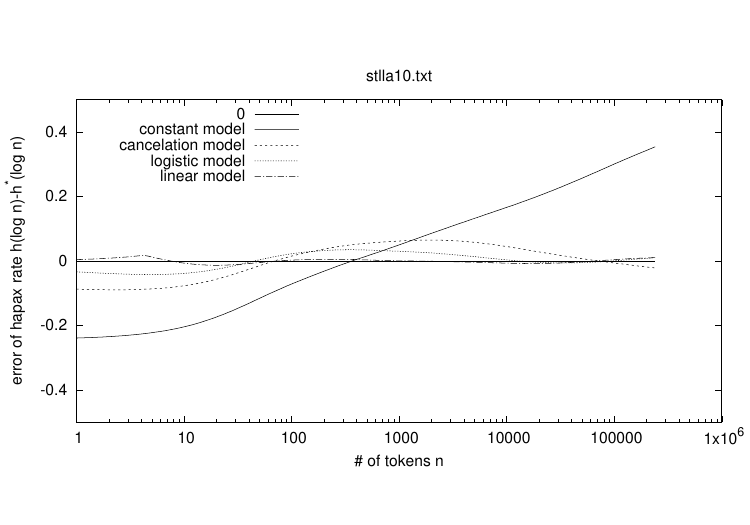}
  \\[-3em]
  \includegraphics[width=0.8\textwidth]{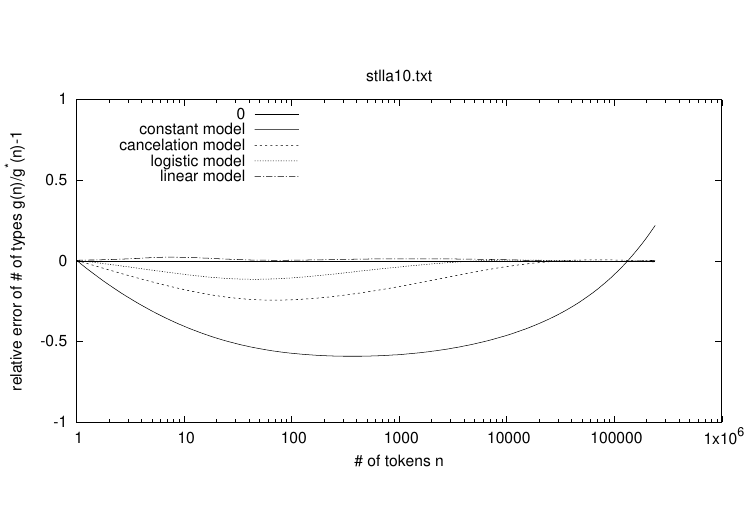}
  \\[-3em]
  \includegraphics[width=0.8\textwidth]{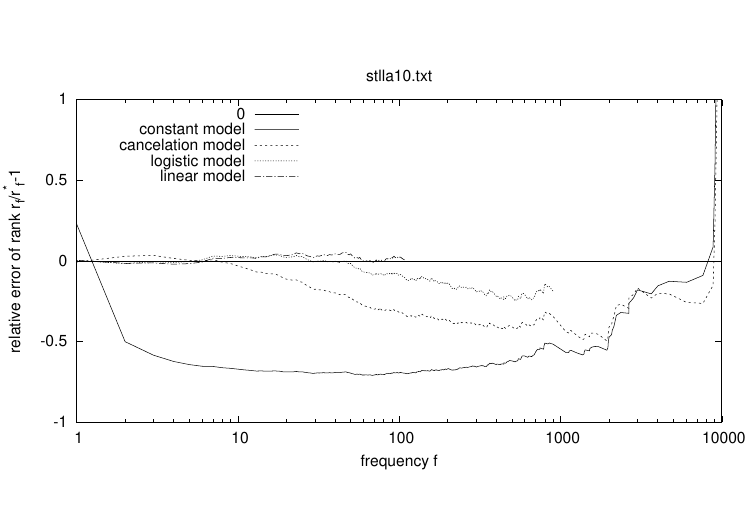}
  \vspace{-2em}
  \caption{J. Swift, \emph{The Journal to Stella}.\label{figstlla10R}}
\end{figure}

\begin{figure}[p]
  \centering
  \vspace{-2em}
  \includegraphics[width=0.8\textwidth]{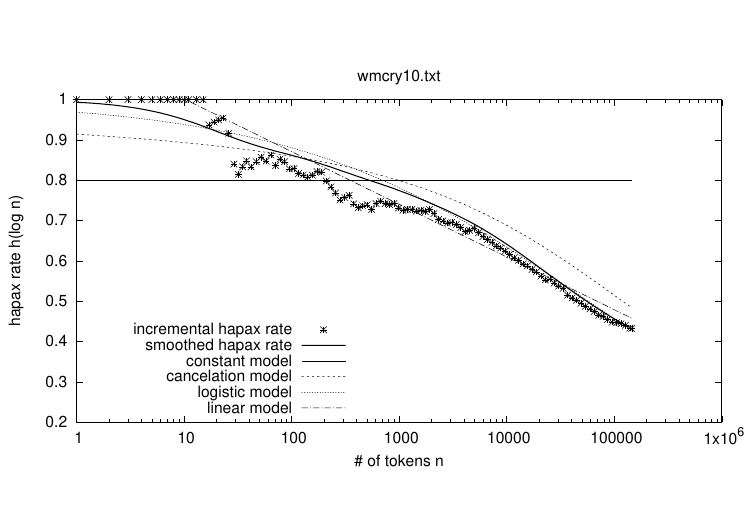}
  \\[-3em]
  \includegraphics[width=0.8\textwidth]{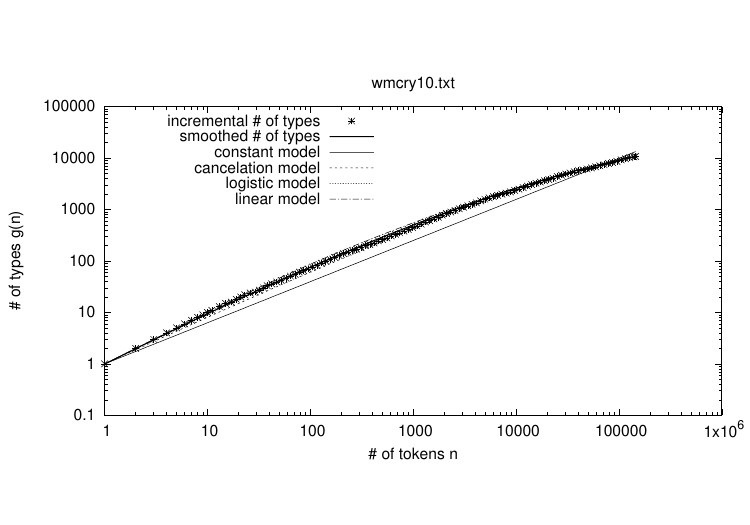}
  \\[-3em]
  \includegraphics[width=0.8\textwidth]{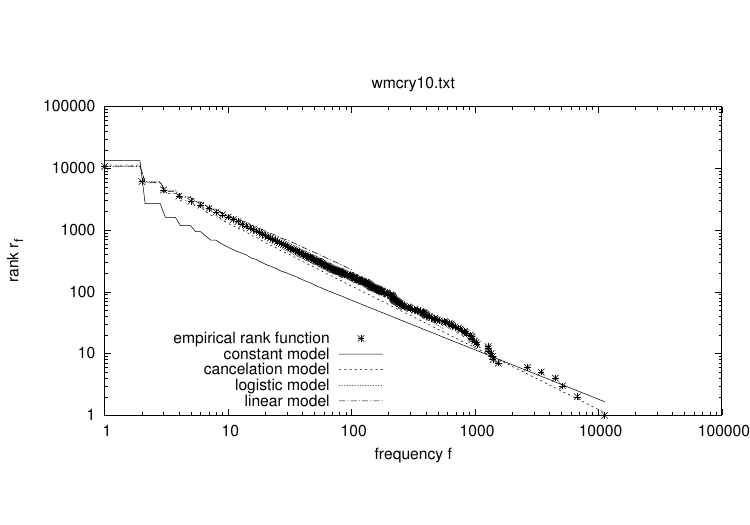}
  \vspace{-2em}
  \caption{G. Smith, \emph{Life of William Carey}.\label{figwmcry10F}}
\end{figure}

\begin{figure}[p]
  \centering
  \vspace{-2em}
  \includegraphics[width=0.8\textwidth]{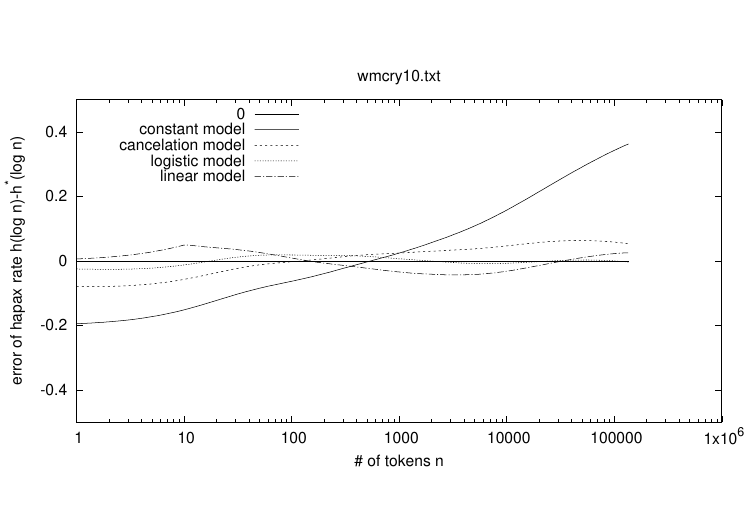}
  \\[-3em]
  \includegraphics[width=0.8\textwidth]{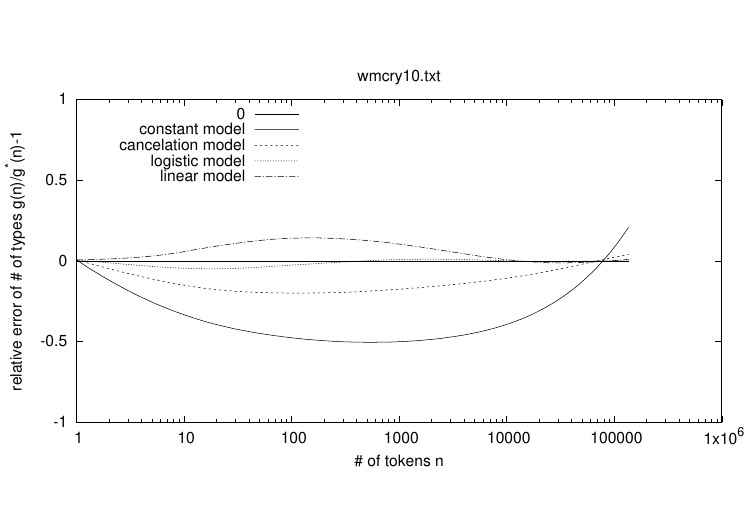}
  \\[-3em]
  \includegraphics[width=0.8\textwidth]{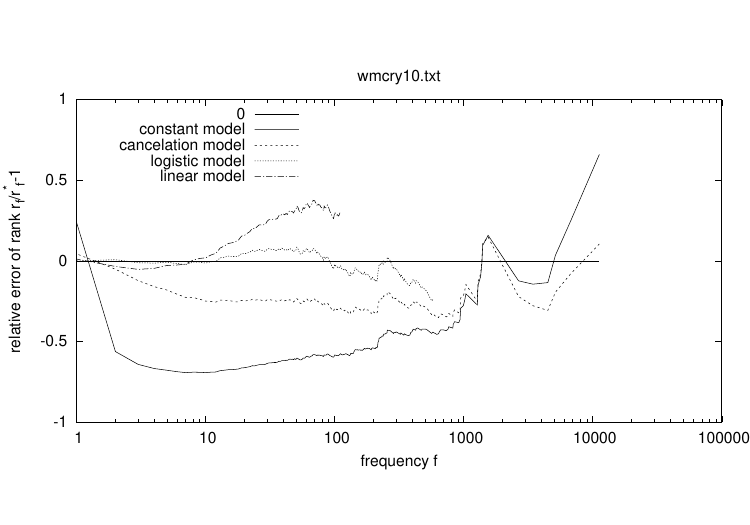}
  \vspace{-2em}
  \caption{G. Smith, \emph{Life of William Carey}.\label{figwmcry10R}}
\end{figure}

For each of the 14 texts investigated in the main article, we produced
three plots depicting: the hapax rate, the vocabulary size (Heaps' law
plot), and the rank function (Zipf's law plot), and three plots
depicting the respective fitting residuals.  We reproduce these plots
as Figures 2 and 3 of the main article and Figures
\ref{fig1ours10F}--\ref{figwmcry10R} of the present report. The legend
to the plots was described in the main article.

In Figures \ref{fig800lg10F} and \ref{figmt7bg10F}, we observe a
surprising piecewise constant plot of the incremental number of
types. This has to do with peculiar but perfectly understandable
beginnings of the corresponding source texts. The beginning of the
first text is:
\\[1em]
{\tt\footnotesize%
  EIGHT HUNDRED LEAGUES ON THE AMAZON BY JULES VERNE PART I THE
  \\
  GIANT RAFT CHAPTER I A CAPTAIN OF THE WOODS P H Y J S L Y D D
  \\
  Q F D Z X G A S G Z Z Q Q E H X G K F N D R X U J U G I O C Y
  \\
  T D X V K S B X H H U Y P O H D V Y R Y M H U H P U Y D K J O
  \\
  X P H E T O Z L S L E T N P M V F F O V P D P A J X H Y Y N O
  \\
  J Y G G A Y M E Q Y N F U Q L N M V L Y F G S U Z M Q I Z T L
  \\
  B Q Q Y U G S Q E U B V N R C R E D G R U Z B L R M X Y U H Q
  \\
  H P Z D R R G C R O H E P Q X U F I V V R P L P H O N T H V D
  \\
  D Q F H Q S N T Z H H H N F E P M Q K Y U U E X K T O G Z G K
  \\
  Y U U M F V I J D Q D P Z J Q S Y K R P L X H X Q R Y M V K L
  \\
  O H H H O T O Z V D K S P P S U V J H D THE MAN WHO HELD IN
  \\
  HIS HAND THE DOCUMENT OF WHICH THIS STRANGE ASSEMBLAGE OF LETTERS%
}
\\[1em]
The beginning of the second text is:
\\[1em]
{\tt\footnotesize%
  MARK TWAIN A BIOGRAPHY BY ALBERT BIGELOW PAINE NOTE SIX
  \\
  INDIVIDUAL VOLUMES OF THIS WORK HAVE BEEN PREVIOUSLY
  \\
  PUBLISHED BY THE GUTENBERG PROJECT IN THE FOLLOWING LIST
  \\
  THIS XMB TEXT FILE INCLUDES ALL SIX D W
  \\
  MARK TWAIN A BIOGRAPHY XXXX XXXX BY ALBERT PAINE MTXBGXX TXT XXXX
  \\
  MARK TWAIN A BIOGRAPHY XXXX XXXX BY ALBERT PAINE MTXBGXX TXT XXXX
  \\
  MARK TWAIN A BIOGRAPHY XXXX XXXX BY ALBERT PAINE MTXBGXX TXT XXXX
  \\
  MARK TWAIN A BIOGRAPHY XXXX XXXX BY ALBERT PAINE MTXBGXX TXT XXXX
  \\
  MARK TWAIN A BIOGRAPHY XXXX XXXX BY ALBERT PAINE MTXBGXX TXT XXXX
  \\
  MARK TWAIN A BIOGRAPHY XXXX XXXX BY ALBERT PAINE MTXBGXX TXT XXXX
  \\
  VOLUME I PART X XXXX XXXX MARK TWAIN A BIOGRAPHY THE PERSONAL AND
  \\
  LITERARY LIFE OF SAMUEL LANGHORNE CLEMENS BY ALBERT BIGELOW PAINE%
}
\\[1em]
As one can see, in the first text, the cryptic message was parsed into
individual letters, which are only 26 types. By contrast, the second
text declares that it consists of six different parts, whose names
were collapsed by mapping all digits onto the same letter X. Both
phenomena caused a local flattening of the incremental vocabulary size
plot.

%%%%%%%%%%%

%%%%%%%%%%%

%%%%%%%%%%%

%%%%%%%%%%%

%%%%%%%%%%%

%%%%%%%%%%%

%%%%%%%%%%%

%%%%%%%%%%%

%%%%%%%%%%%

%%%%%%%%%%%

%%%%%%%%%%%

%%%%%%%%%%%

%%%%%%%%%%%

\end{document}